\documentclass[]{digital_agent_hub}

\usepackage{pifont}
\usepackage{float}
\usepackage{longtable}
\usepackage{listings}
\usepackage{tabularx}
\usepackage{amssymb}
\usepackage{makecell}
\usepackage{xurl}
\usepackage{makecell}
\usepackage{graphicx}
\usepackage{booktabs}
\usepackage{pifont}
\usepackage[table]{xcolor}
\usepackage{hyperref}
\usepackage{algorithm}
\usepackage{paralist}
\usepackage{algpseudocode}
\usepackage{xcolor}
\usepackage{multirow}
\usepackage{wrapfig}
\usepackage{xspace}
\usepackage{color}
\usepackage{multirow}
\usepackage{ifthen}

\long\def\ignorethis#1{}

\definecolor{gray}{rgb}{0.35,0.35,0.35}
\definecolor{MyBlue}{rgb}{0,0.2,0.8}
\definecolor{MyRed}{rgb}{0.8,0.2,0}
\definecolor{MyGreen}{rgb}{0.0,0.5,0.1}
\definecolor{MyGray}{rgb}{0.4,0.4,0.4}

\newlength\paramargin
\newlength\figmargin
\newlength\subfigmargin
\newlength\secmargin
\newlength\subsecmargin
\newlength\tabmargin
\newlength\eqmargin

\usepackage{array}
\newcolumntype{L}[1]{>{\raggedright\let\newline\\\arraybackslash\hspace{0pt}}m{#1}}
\newcolumntype{C}[1]{>{\centering\let\newline\\\arraybackslash\hspace{0pt}}m{#1}}
\newcolumntype{R}[1]{>{\raggedleft\let\newline\\\arraybackslash\hspace{0pt}}m{#1}}

\def\eg{e.g.,~}

\newcommand{\figref}[1]{Fig.~\ref{fig:#1}}
\newcommand{\tabref}[1]{Table~\ref{tab:#1}}

\newcommand{\Paragraph}[1]{\noindent\textbf{#1}}

\definecolor{headerblue}{HTML}{E8F0FA}
\definecolor{bggray}{HTML}{F2F2F2}
\definecolor{textgray}{HTML}{666666}
\definecolor{tabletwored}{HTML}{F02F1D}
\definecolor{tabletwogreen}{HTML}{3CA52C}

\newcommand{\tabletwocmark}{\textcolor{tabletwogreen}{\ding{51}}}
\newcommand{\tabletwoxmark}{\textcolor{tabletwored}{\ding{55}}}
\newcommand{\cutverse}{{\sffamily\bfseries CutVerse}}

\newcommand{\gameid}[1]{\begingroup\urlstyle{tt}\nolinkurl{#1}\endgroup}
\newcommand{\taxonomymodellogo}[2]{\raisebox{-0.05em}{\includegraphics[height=0.9em]{figs/logos/#1}} #2}

\definecolor{headerblue}{HTML}{eef2f8}
\definecolor{bggray}{HTML}{F2F2F2}
\definecolor{textgray}{HTML}{4A4A4A}
\definecolor{rankbgyellow}{HTML}{FFF8E1}
\definecolor{projectpagelink}{HTML}{DA2F8A}

\newcommand{\captfont}[1]{#1}

\begin{document}

\title{\cutverse{}: A Compositional GUI Agents Benchmark for Media Post-Production Editing}

\author{Haobo Hu$^{*}$$^{1}$, Xiangwu Guo$^{*}$$^{2}$, Zhiheng Chen$^{2}$, Difei Gao$^{3}$, Haotian Liu$^{1}$,Libiao Jin$^{1}$,Qi Mao$^{1}$$^{\dag}$}
\affiliation{$^{1}$MIPG, Communication University of China, $^{2}$National University of Singapore,$^{3}$USEIT AI}

\projectpage{\href{https://github.com/CUC-MIPG/CutVerse}{\textcolor{projectpagelink}{https://github.com/CUC-MIPG/CutVerse}}}
\firstpagefootnote{$^{*}$Equal contribution. $^{\dag}$Corresponding authors.}

\abstract{
While GUI agents have made significant progress in web navigation and basic operating system tasks, their capabilities in professional creative workflows remain largely underexplored. To bridge this gap, we introduce \textbf{\cutverse{}}, a benchmark designed to systematically evaluate autonomous GUI agents in realistic media post-production environments. We curate expert demonstrations across $7$ professional applications (e.g., Premiere Pro, Photoshop), covering $186$ complex, long-horizon tasks grounded in authentic editing workflows, involving dense multimodal interfaces and tightly coupled interaction sequences. To support scalable evaluation, we develop a lightweight parser that transforms raw screen recordings and low-level interaction logs into structured, compositional GUI action trajectories with precise grounding.
Extensive evaluations reveal that existing agents achieve only 36.0\% task success on realistic media editing tasks, underscoring the challenges posed by complex, long-horizon media post-production workflows in our benchmark.While current models demonstrate promising spatial grounding, multimodal alignment, and coordinated action execution, they remain limited in long-horizon reliability and domain-specific planning.
}
\maketitle
\begin{figure}[H]
\vspace{-4mm}
  \includegraphics[width=\textwidth]{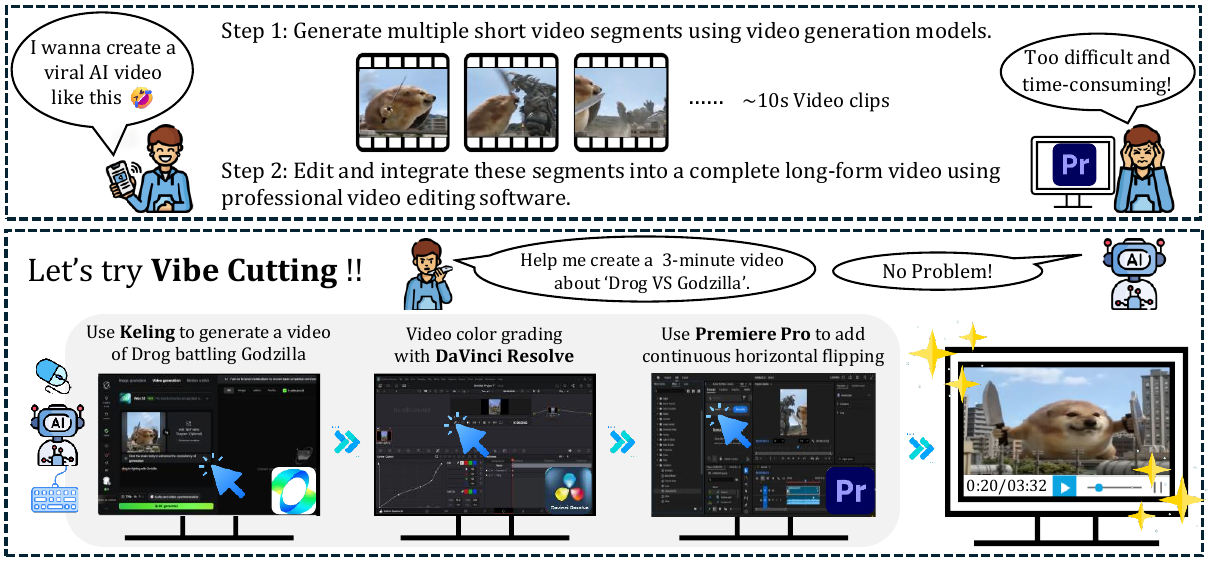}
\vspace{-6mm}
\caption{\textbf{CutVerse: A benchmark for evaluating GUI agents in media post-production.} 
\textbf{Top:} Existing AI video creation pipelines require manual composition of generated clips within professional editing software. 
\textbf{Bottom:} CutVerse evaluates GUI agents on realistic post-production tasks across diverse professional tools, covering complete workflows such as timeline editing, visual effects, audio alignment, and content composition through real software interaction.}
  \label{fig:teaser}
\end{figure}

\makeabstract

\setcounter{tocdepth}{2}
\tableofcontents
\clearpage

\section{Introduction}
\label{sec:intro}

The development of computer-use agents (CUA)\cite{hu2024dawnguiagentpreliminary,hu2025osagentssurveymllmbased} emerges as a promising direction for bridging natural language instructions with executable actions in software environments. 
By leveraging vision-language models\cite{hong2024cogagentvisuallanguagemodel,cheng2024seeclickharnessingguigrounding}, these agents can perceive screen content\cite{lu2024omniparserpurevisionbased,yang2023setofmarkpromptingunleashesextraordinary} and generate coherent interaction sequences\cite{xu2025aguvisunifiedpurevision}, enabling automation across a wide range of web and desktop applications. 
Recent advances demonstrate strong capabilities in structured tasks, including web navigation~\cite{xu2025agenttrek,zhou2024webarenarealisticwebenvironment}, official software operation~\cite{xie2024osworld,bonatti2025windows}, and basic system-level interactions~\cite{wang2025opencua,kapoor2024omniact}, marking an important step toward general-purpose computer-use agents\cite{wu2024osatlasfoundationactionmodel}. 
As agents master these general-purpose domains, their capability boundaries remain fundamentally underexplored when confronted with the intricate, unstructured demands of highly professional real-world workflows.

A representative yet underexplored domain is media post-production. 
Compared to existing scenarios, professional creative software presents substantially higher interface density\cite{zhao2026worldguiinteractivebenchmarkdesktop}, more fine-grained and intricate interaction patterns, and significantly longer execution horizons. 
Users must orchestrate a sequence of tightly coupled operations, including timeline manipulation, layer composition, parameter tuning, and cross-modal alignment between audio and visual signals. 
Such workflows impose strong requirements on spatial precision, temporal consistency, and coordinated multi-modal control, posing fundamental challenges that are not captured by current evaluation settings.

However, evaluating CUA agents in media post-production further introduces substantial system-level and infrastructural challenges. 
Unlike conventional benchmarks, which operate in lightweight and relatively stable environments, media editing workflows involve significantly higher memory footprints, complex and continuously evolving system states, and substantially more diverse and longer action trajectories. 
These characteristics place strict demands on environment reproducibility, state management, and execution stability. Existing benchmarks and datasets are not designed to support such high-fidelity, resource-intensive scenarios, making it difficult to reliably instantiate and evaluate agent behavior in realistic media production settings.

These limitations highlight the need for an evaluation framework that captures the complexity of real-world creative workflows, including continuous GUI interaction, multimodal perception, and long-horizon execution. 
To address these challenges, we introduce \textbf{CutVerse}, a benchmark designed to systematically evaluate CUA agents in realistic media post-production environments. We further build a robust infrastructure that includes (i) \textbf{a lightweight parser} that transforms raw multimodal interaction logs into structured GUI trajectories with grounding annotations, and (ii) a Windows-based virtual environment that enables agents to execute actions directly within software to support scalable and reproducible evaluation.

In parallel, AIGC-based pipelines primarily target high-level semantic alignment and visual consistency~\cite{huang2025filmasterbridgingcinematicprinciples,11092919,he2025dreamstoryopendomainstoryvisualization}, while code-driven approaches are often limited to simple operations such as direct video stitching. 
\begin{wrapfigure}{r}{0.45\textwidth}
  \vspace{-3mm} 
  \centering
  \includegraphics[width=0.95\linewidth]{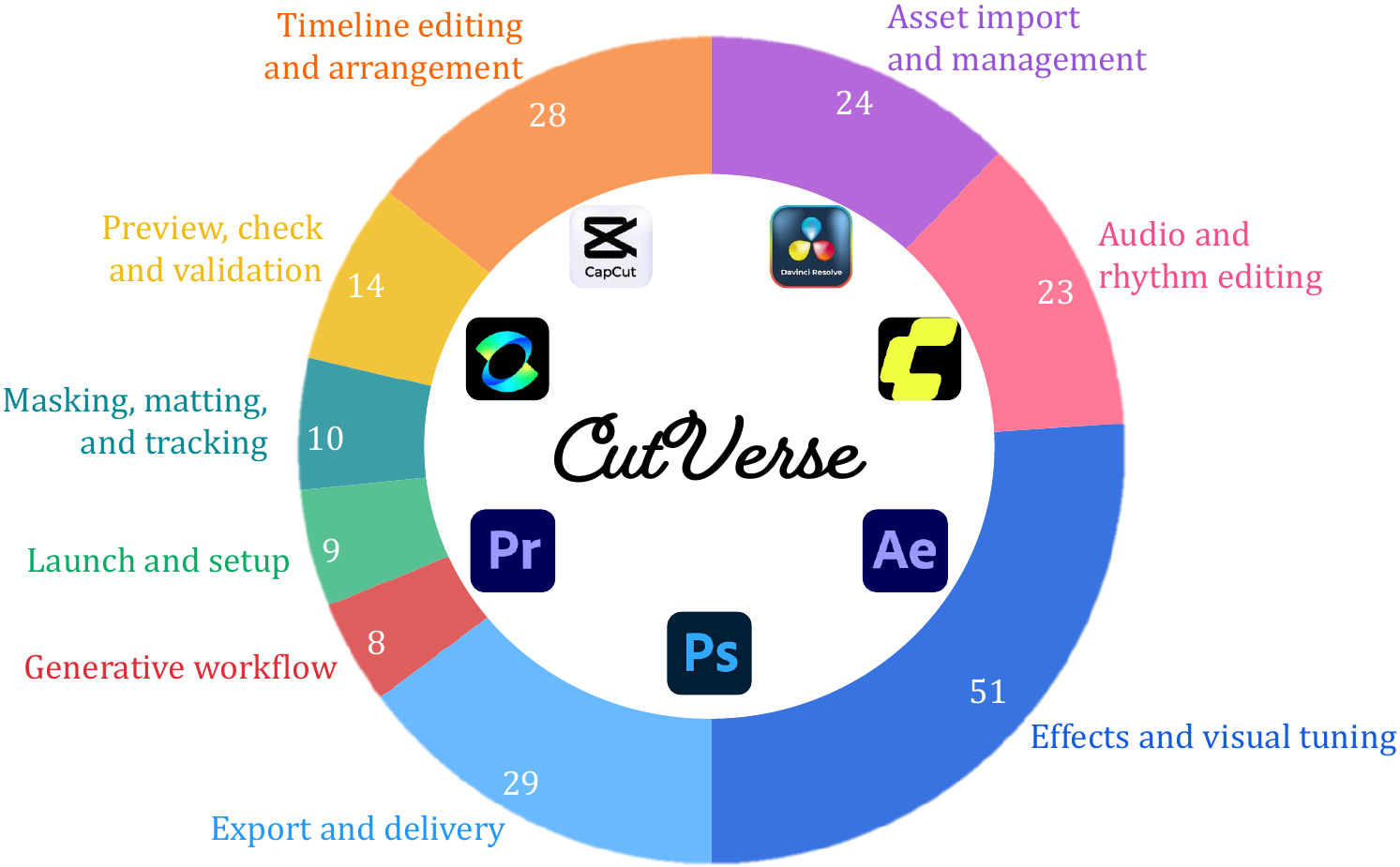}
  \caption{\textbf{CutVerse task and software ecosystem.} The inner circle displays integrated post-production applications. The outer ring categorizes 186 human-verified tasks across nine domains.}
  \label{fig:task_distribution}
  \vspace{-3mm} 
\end{wrapfigure}
Both paradigms struggle to support fine-grained editing under fixed source content, including layer-wise color grading, geometric transformations, and precise transition effects that are fundamental to professional post-production. 
To bridge this gap, our benchmark is grounded in complete, real-world media post-production workflows, comprising 186 well-designed tasks across 7 professional software platforms, each paired with a specific virtual machine checkpoint and manually recorded interaction trajectories to faithfully capture authentic editing processes for realistic agent evaluation.

Extensive experiments reveal a substantial performance gap. 
Even the strongest models struggle with sustained execution in complex workflows, exhibiting failures in spatial grounding, temporal coordination, and compositional interaction. 
These results suggest that current agents, while effective in simplified domains, remain far from reliable deployment in professional creative environments.
Beyond benchmarking, our findings point toward a broader paradigm for AI-assisted media production, which we term \textit{Vibe Cutting}, where generation provides multimodal assets and agents transform them into structured outputs through real software interaction, as illustrated in \figref{teaser}. 
As a broader vision, we anticipate that CutVerse will provide a practical foundation for advancing end-to-end multimedia production.

Our contributions are summarized as follows:
\begin{itemize}
\item We introduce \textbf{CutVerse}, a comprehensive dataset comprising 186 complex, long-horizon tasks across 7 professional applications, specifically targeting realistic media post-production workflows.
    \item  We build an end-to-end pipeline consisting of a infrastructure parser that converts raw multimodal logs into structured GUI trajectories, and a Windows VM-based evaluation environment for authentic agent execution.
    \item  We design fine-grained evaluation metrics that move beyond traditional Success Rates (SR) to strictly reflect the fine-grained operations and specific characteristics of creative applications.
    \item  Extensive evaluations of state-of-the-art VLMs reveal a striking performance gap, exposing critical bottlenecks in handling spatially dense layouts and compositional GUI actions.
\end{itemize}

\section{Related Work}
\label{sec:related}

\subsection{AIGC Agents}
Recent AIGC agents leverage planner-executor paradigms~\cite{wei2022chain,yao2022react} and tool augmentation~\cite{schick2023toolformer} to automate multimodal content generation~\cite{Wang2024LAVELA,wang2024genartist,li2024anim,shi2025animaker,zheng2024videogen,huang2025filmasterbridgingcinematicprinciples,zhang2026stagestoryboardanchoredgenerationcinematic,11092919,he2025dreamstoryopendomainstoryvisualization}. However, these frameworks predominantly target coarse-grained semantic alignment and high-level visual consistency. When confronted with the rigorous demands of professional multimedia post-production, including fine-grained video effects (VFX), precise timeline manipulations, and complex transition editing, existing AIGC architectures prove fundamentally inadequate. They currently lack the execution granularity required to navigate the intricate, trivial operational workflows essential for professional-grade media post-production.

\subsection{GUI Agents and Benchmarks}
While recent VLM-based GUI agents~\cite{hong2024cogagent,xue2026evocuaevolvingcomputeruse,lin2025showui,qin2025ui,chen2025uiinsenhancingguigrounding,gu2025uivenustechnicalreportbuilding,li2025screenspotpro,xu2025aguvis,zhang2025tonguiinternetscaletrajectoriesmultimodal,ui-tars-15-seed} exhibit strong interactive capabilities across general-purpose domains~\cite{nguyen-etal-2025-gui,gao2024assistguitaskorienteddesktopgraphical,lu2025guiodyssey,rawles2023androidwildlargescaledataset,kong2025mobileworldbenchmarkingautonomousmobile} like web navigation~\cite{deng2023mind2web,xu2025agenttrek,kapoor2024omniact,zhou2024webarenarealisticwebenvironment,koh-etal-2024-visualwebarena} and operating systems~\cite{xie2024osworld,yang2025macosworld,bonatti2025windows,liu2026scalecua,lin2024videogui,wang2025opencua,nayak2025uivision,rawles2025androidworld,10.5555/3666122.3667612}, they aim to bridge natural language instructions and executable actions within interactive software environments.  
However, the specialized domain of media post-production remains severely underexplored. Professional editing environments present unique challenges characterized by exceptionally dense interface layouts and long-horizon operational sequences. Because existing GUI benchmarks are largely constrained to simplified and short-step interactions, they are incapable of effectively evaluating the complex, multi-step execution trajectories inherent to real-world editing workflows.

\begin{figure*}[!t]
    \centering
    \includegraphics[width=1.\textwidth]{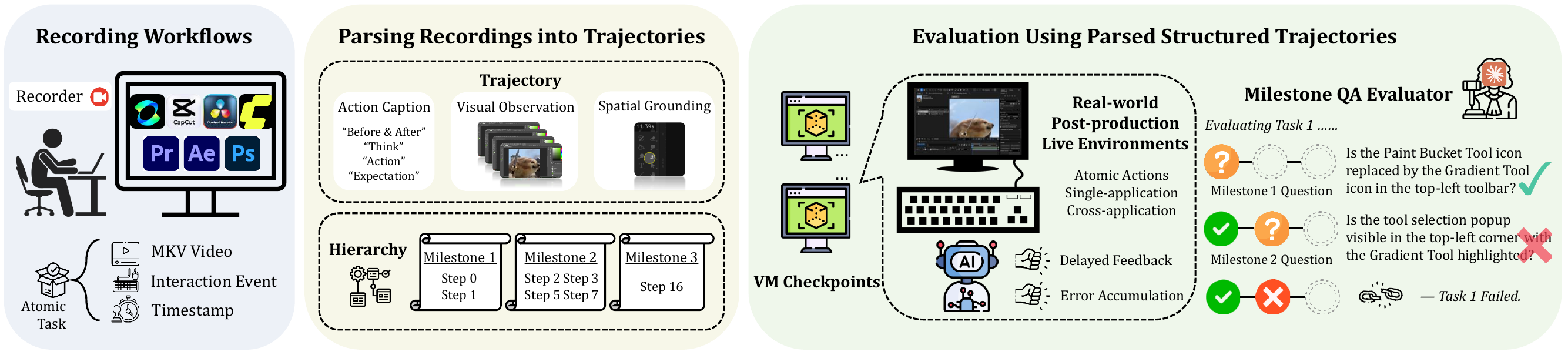}
    \caption{
\textbf{The CutVerse data and evaluation pipeline.} (1) \textbf{Recording}: Capturing synchronized expert workflows across professional applications. (2) \textbf{Parsing}: Structuring raw data into milestone-driven trajectories with rich spatiotemporal grounding. (3) \textbf{Evaluation}: Assessing full agent trajectories in live environments via a post-hoc Milestone QA Evaluator. Although the entire task is executed, intermediate milestone failures still dictate overall task failure, authentically exposing error accumulation in long-horizon editing.
    }
    \label{fig:framework}
\end{figure*}

\subsection{Media Creative Benchmarks}
Existing media creative benchmarks~\cite{huang2023vbench,huang2025vbench++,liu2025shotbench,chen2026ivebench,zheng2025cmlbench,huang2024comfybench,liang2023editval,zhuang2025vistorybench} have driven significant advancements in assessing the high-dimensional perceptual quality and semantic fidelity of generated multimodal content. Nevertheless, these evaluations remain fundamentally output-oriented. There is a critical absence of standardized protocols capable of comprehensively evaluating the interaction density of professional creative tools, specifically the precise cutting actions and dynamic effect tuning executed during the creation process. To address this gap, \textbf{CutVerse} introduces a rigorous evaluation standard that shifts the focus from static output assessment to the dynamic, trajectory-based verification of professional media manipulation.

\begin{table*}[!t]
\centering
\caption{
\captfont{
\textbf{Comparison of GUI agent benchmarks across platforms and workflow complexity.}
\textbf{Media Tier} denotes the professional level of supported multimedia creation environments, categorized into \textit{Basic} (lightweight/consumer-grade tools) and \textit{Pro} (paid professional post-production software, e.g., Adobe Premiere Pro and After Effects);
\textbf{E2E Workflow} indicates the inclusion of compositional, long-horizon editing workflows that culminate in a final exported video product, as opposed to isolated atomic kills;
\textbf{AIGC} indicates the integration of generative AI tools and pipelines;
\textbf{Human} refers to human-curated data;
\textbf{Env.} indicates the availability of live, executable environments for interactive evaluation.
}
}
\vspace{-3mm}
\label{tab:comparison_vm}
\scriptsize
\setlength{\tabcolsep}{4pt}
\renewcommand{\arraystretch}{1.2}
\resizebox{\textwidth}{!}{
\begin{tabular}{llcccccc}
\toprule
\rowcolor{headerblue}
\textbf{Benchmark} &
\textbf{Platform} &
\thead{Tasks} &
\thead{Media\\Tier} &
\thead{E2E\\Workflow} &
\textbf{AIGC} &
\textbf{Human} &
\textbf{Env.} \\
\midrule

\rowcolor{bggray}
\multicolumn{8}{c}{\textcolor{textgray}{\textbf{\textit{Web \& Mobile}}}} \\
\midrule
Mind2Web~\cite{deng2023mind2web}    & Web     & 2,350  & \tabletwoxmark & \tabletwoxmark & \tabletwoxmark & \tabletwocmark & \tabletwoxmark \\
AgentTrek~\cite{xu2025agenttrek}   & Web     & 10,398 & \tabletwoxmark & \tabletwoxmark & \tabletwoxmark & \tabletwoxmark & \tabletwoxmark \\
AITW~\cite{rawles2023androidwildlargescaledataset}        & Mobile  & 715K   & \tabletwoxmark & \tabletwoxmark & \tabletwoxmark & Mix.   & \tabletwoxmark \\
GUI-Odyssey~\cite{lu2025guiodyssey} & Mobile  & 8,334  & \tabletwoxmark & \tabletwoxmark & \tabletwoxmark & Mix.   & \tabletwoxmark \\

\midrule
\rowcolor{bggray}
\multicolumn{8}{c}{\textcolor{textgray}{\textbf{\textit{Desktop \& General}}}} \\
\midrule
OmniACT~\cite{kapoor2024omniact}   & Desktop + Web & 9,802  & \tabletwoxmark & \tabletwoxmark & \tabletwoxmark & \tabletwocmark & Partial \\
OpenCUA~\cite{wang2025opencua}   & Desktop       & 22,625 & Basic  & \tabletwoxmark & \tabletwoxmark & \tabletwocmark & \tabletwoxmark  \\
VideoGUI~\cite{lin2024videogui}  & Desktop       & 178    & Pro    & \tabletwoxmark & \tabletwoxmark & Mix.   & \tabletwoxmark  \\
ScaleCUA~\cite{liu2026scalecua}  & Cross-platform& $\sim$19K & Basic & \tabletwoxmark & \tabletwoxmark & Mix. & \tabletwoxmark  \\
Window Agent Arena~\cite{bonatti2025windows}  & Desktop &  154 & \tabletwoxmark & \tabletwoxmark & \tabletwoxmark & Mix. & \tabletwocmark  \\
OSWorld~\cite{xie2024osworld}   & Desktop       & 369    & \tabletwoxmark & \tabletwoxmark & \tabletwoxmark & \tabletwocmark & \tabletwocmark  \\
macOSWorld~\cite{yang2025macosworld} & Desktop  & 230    & \tabletwoxmark & \tabletwoxmark & \tabletwoxmark & \tabletwocmark & \tabletwocmark \\

\midrule
\textbf{CutVerse (Ours)} 
& \textbf{Desktop} 
& \textbf{186} 
& \textbf{Pro} 
& \tabletwocmark 
& \tabletwocmark 
& \tabletwocmark 
& \tabletwocmark \\
\bottomrule
\end{tabular}
}
\end{table*}

\section{The CutVerse Benchmark}
\label{sec:benchmark}

To systematically evaluate GUI agents in realistic creative workflows, we introduce \textbf{CutVerse}, a comprehensive benchmark engineered to bridge the critical ``last mile'' between isolated artificial intelligence-generated content (AIGC) and production-ready media. To encapsulate real-world complexities into a highly scalable evaluation infrastructure, CutVerse formulates an end-to-end pipeline comprising high-fidelity data recording, structural multimodal parsing, and systematic dual-mode evaluation.



\subsection{Task Formulation}

Unlike conventional benchmarks targeting static webpages, agents in CutVerse are immersed in dynamic workspaces characterized by severe multimodal information overload and high resource complexity. Professional editing necessitates managing extensive asset libraries, continuous audio waveforms, and dense parameter panels simultaneously. Crucially, successful navigation in this environment demands a dual-tiered visual perception capability: beyond basic UI widget localization, agents must exhibit profound media content comprehension. They are required to interpret the semantic, aesthetic, and temporal nuances of the underlying visual and auditory streams to formulate contextually appropriate editing decisions.

Furthermore, CutVerse redefines the benchmark task as a holistic objective requiring rigorous cross-modal alignment and spatiotemporal synchronization—such as precisely aligning an audio effect to a specific dynamic event within a video frame. To mirror modern creative paradigms, our tasks mandate seamless cross-application workflows, dictating that agents fluently orchestrate operations from generating raw visual assets via AIGC node-based interfaces (e.g., ComfyUI) to refining and composing them within traditional nonlinear editing platforms (e.g., Adobe Premiere Pro).

To faithfully reflect these steep technical barriers, CutVerse enforces an \textit{anthropomorphic action space} entirely grounded in vision-only perception. Rather than executing privileged software APIs, agents are compelled to embody the human motor-cognitive loop. They interact with the software exclusively through continuous mouse drag-and-drops, precise coordinate-based clicks, and complex keyboard shortcut combinations. In stark contrast to web tasks driven by structured HTML DOM trees, media post-production relies heavily on unstructured multi-track timelines. Executing tasks within this space necessitates rigorous pixel-level grounding, compelling agents to translate high-level creative intentions into concrete physical operations precisely as a human creator would.

\begin{table*}[!t]
    \centering
    \caption{\textbf{Task-centric analysis of CutVerse.} We reorganize workflows by task type, combining distribution statistics with interaction complexity and functional coverage.}
    \vspace{-3mm}
    \label{tab:task_combined}

    \definecolor{checkgreen}{HTML}{2E7D32} 
    \newcommand{\gcheck}{\textcolor{checkgreen}{\ding{51}}}

    \scriptsize
    \setlength{\tabcolsep}{3pt}
    \renewcommand{\arraystretch}{1.3} 

    \resizebox{\linewidth}{!}{
    \begin{tabular}{llcccccccccccc}
    \toprule
    \rowcolor{headerblue}
    & & & & & & & \multicolumn{7}{c}{\textbf{Functional Coverage}} \\
    \rowcolor{headerblue}
    \multirow{-2}{*}{\textbf{Task Type}} & 
    \multirow{-2}{*}{\textbf{Primary Software}} & 
    \multirow{-2}{*}{\textbf{Count}} & 
    \multirow{-2}{*}{\textbf{Ratio}} & 
    \multirow{-2}{*}{\thead{Avg.\\Duration}} & 
    \multirow{-2}{*}{\thead{Avg.\\Steps}} & 
    \multirow{-2}{*}{\textbf{Complexity}} & 
    \textbf{Edit} & \textbf{Audio} & \textbf{VFX} & \textbf{Motion} & \textbf{Color} & \textbf{AIGC} & \textbf{Asset} \\
    \midrule

    Effects and visual tuning & After Effects / Photoshop & 51 & 27.4\% & 52.81 & 20.27 & Extreme & \gcheck &  & \gcheck & \gcheck & \gcheck &  &  \\
    Export and delivery       & All Platforms             & 29 & 15.6\% & 48.98 & 15.41 & High    &  & \gcheck &  &  &  &  & \gcheck \\
    Asset import and management & All Platforms           & 24 & 12.9\% & 43.01 & 20.83 & High    & \gcheck &  &  &  &  &  & \gcheck \\
    Audio and rhythm editing  & Premiere Pro / JianYing   & 23 & 12.4\% & 45.94 & 26.00 & High    & \gcheck & \gcheck &  &  &  &  &  \\
    Timeline editing and arrangement & Premiere Pro / DaVinci & 18 & 9.7\% & 46.48 & 23.67 & High & \gcheck & \gcheck & \gcheck &  &  &  &  \\
    Preview, check, and validation & All Platforms        & 14 & 7.5\%  & 22.01 & 5.50  & Medium  & \gcheck &  &  &  &  &  & \gcheck \\
    Masking, matting, and tracking & After Effects / Photoshop & 10 & 5.4\% & 72.98 & 25.40 & Extreme & \gcheck &  & \gcheck & \gcheck & \gcheck &  &  \\
    Launch and setup          & All Platforms             & 9  & 4.8\%  & 31.18 & 7.56  & Low     &  &  &  &  &  &  & \gcheck \\
    Generative workflow       & ComfyUI / Keling          & 8  & 4.3\%  & 35.45 & 10.00 & Medium  &  &  &  &  &  & \gcheck & \gcheck \\

    \bottomrule
    \end{tabular}
    }
    \vspace{-4mm}
\end{table*}

\subsection{Scalable Evaluation Infrastructure and Capability Decomposition}
To rigorously support the dynamic and human-centric nature of media post-production, CutVerse is instantiated upon a robust, scalable evaluation infrastructure powered by a custom Windows virtualization engine. Crucially, to genuinely evaluate agents in authentic scenarios, the environment enforces a strict human-aligned execution paradigm. Rather than relying on privileged backend APIs—which are largely nonexistent in professional creative suites—the engine isolates each task within a resettable virtual machine, restricting agent interactions entirely to simulated, low-level mouse and keyboard events driven by live visual feedback. This architectural design forces autonomous agents to operate under the exact systemic and cognitive constraints as human professionals. Furthermore, precise state checkpointing guarantees systemic reproducibility and visual consistency across large-scale evaluations, ensuring a reliable testbed without compromising the live, interactive nature of the host operating system.

Operating synergistically with this execution engine is a dedicated multimodal parsing pipeline, designed to transform unstructured human demonstrations into evaluable, structured formats. Specifically, the parser meticulously synchronizes high-framerate screen recordings with low-level I/O event logs, extracting spatiotemporally aligned action sequences. Through this rigorous alignment, continuous expert workflows—comprising natural language instructions, raw video frames, and complex keystrokes—are translated into structured multimodal trajectories. Each discrete step is firmly grounded in its corresponding visual state and semantic context, thereby effectively bridging the semantic gap between continuous pixel arrays and actionable agent representations.

Beyond simple trajectory extraction, this parsing infrastructure introduces a profound paradigm shift by decomposing long-horizon, monolithic workflows into hierarchical semantic milestones, which we subsequently map to \textit{transferable atomic capabilities}. While specific post-production tasks exhibit massive combinatorial diversity (e.g., color grading a cinematic shot versus synthesizing a dynamic transition), their underlying milestones rely on a finite set of foundational, cross-domain skills. These include temporal navigation across multi-track timelines, granular parameter fine-tuning, and cross-modal asset retrieval. By decoupling complex tasks into these atomic capabilities, CutVerse achieves an unprecedented level of fine-grained diagnostic resolution. It transcends binary task success rates, empowering researchers to precisely quantify whether an agent has acquired generalizable editing skills that can seamlessly transfer across disparate generative tools and traditional software ecosystems.

\subsection{Dataset Construction and Statistical Complexity}
\label{sec:generalist_agents}

Constructed atop our robust virtualization infrastructure, the CutVerse dataset encapsulates the authentic complexity of modern media workflows through 2.43 hours of high-fidelity recording. As illustrated in Figure~\ref{fig:task_distribution} and detailed in Table~\ref{tab:task_combined}, it yields 186 human-verified tasks and 3,484 atomic GUI interactions (averaging 23.8 interactions per minute) spanning nine functional domains. Beyond sheer scale, CutVerse covers the entire production pipeline—from procedural asset management to complex visual tuning. Crucially, we push the benchmark beyond traditional industry-standard software (e.g., Adobe Premiere Pro, After Effects) by incorporating interactions with emerging generative platforms like Keling, Jimeng, and ComfyUI. This hybrid composition effectively mirrors contemporary creative paradigms, where users fluidly orchestrate end-to-end workflows by synthesizing raw AIGC materials and subsequently refining them within conventional editing ecosystems.

A defining characteristic of CutVerse is its pronounced long-horizon complexity, which stringently evaluates an agent's capacity for sustained planning and continuous multimodal context maintenance. The dataset exhibits a severe long-tail distribution, averaging 18.73 steps per trajectory—substantially surpassing standard web-navigation benchmarks—with peak execution horizons reaching 239 steps. To systematically quantify this, we stratify workflows along a complexity spectrum from \textit{Low} to \textit{Extreme} (see Table~\ref{tab:task_combined}). Notably, tasks demanding dense cross-modal alignment and fine-grained spatiotemporal precision—such as audio rhythm editing (averaging 26.00 steps) or masking and tracking (25.40 steps)—exhibit exceptionally high interaction density. This challenge is further exacerbated across software boundaries, where cross-application workflows escalate to an average of 21.20 steps compared to 17.56 for isolated applications. Consequently, mastering CutVerse necessitates reliably executing prolonged sequences of compositional operations without losing the overarching semantic goal.

Finally, our statistical analysis exposes the exceptional visual parsing threshold inherent to professional creative environments. A breakdown of interacting UI elements reveals that timelines completely dominate the visual focus, accounting for 46.07\% of all operations, followed closely by complex layer and track controls at 25.32\%. In stark contrast to standard web DOMs populated by discrete HTML buttons, timelines operate as unstructured, spatiotemporal interfaces demanding granular spatial adjustment and continuous coordination (e.g., continuous drag-and-drop, precise multi-key combos). The overwhelming prevalence of these elements definitively shifts the evaluation bottleneck from simple point-and-click navigation to maintaining pixel-level audio-visual grounding amidst severe multimodal information overload.

\subsection{Online Execution and Automated Milestone Assessment}

\begin{table*}[!t]
\centering
\caption{Unified task and milestone success rates across operation categories. Models demonstrate strong capabilities in procedural setup and basic file management, such as generative workflows, software launching, and exporting. However, performance degrades significantly when executing core media editing tasks. The stark contrast between local milestone success and overall task success highlights a fundamental weakness in complex content manipulation, audio coordination, and precise visual tuning.}
\vspace{-3mm}
\label{tab:unified_analysis_task_milestone}

\footnotesize
\setlength{\tabcolsep}{2pt} 
\renewcommand{\arraystretch}{1.15}

\definecolor{rowgreen}{HTML}{A2CF87}
\definecolor{roworange}{HTML}{FDD9B9}
\definecolor{rowblue}{HTML}{D4E6F1} 

\resizebox{\linewidth}{!}{
\begin{tabular}{l | ccccc | ccccc}
\toprule

\multicolumn{1}{l}{\multirow{2}{*}{\textbf{Task Category}}} 
& \multicolumn{5}{c}{\textbf{Task Success Rate}} 
& \multicolumn{5}{c}{\textbf{Milestone Success Rate}} \\
\cmidrule(lr){2-6} \cmidrule(lr){7-11} 
\multicolumn{1}{l}{} 
& \begin{tabular}{@{}c@{}}\taxonomymodellogo{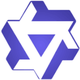}Qwen3\\-32B-T~\cite{yang2025qwen3technicalreport}\end{tabular} 
& \begin{tabular}{@{}c@{}}\taxonomymodellogo{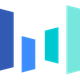}UI-TARS\\-1.5-7B~\cite{qin2025ui}\end{tabular} 
& \begin{tabular}{@{}c@{}}\taxonomymodellogo{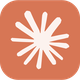}Claude\\-Opus-4.6~\cite{anthropic2026claude46}\end{tabular} 
& \begin{tabular}{@{}c@{}}\taxonomymodellogo{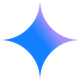}Gemini3\\-flash~\cite{gemini3_2026}\end{tabular} 
& \begin{tabular}{@{}c@{}}\taxonomymodellogo{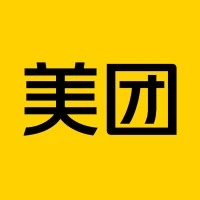}EvoCUA\\-32B~\cite{xue2026evocuaevolvingcomputeruse}\end{tabular} 
& \begin{tabular}{@{}c@{}}\taxonomymodellogo{qwen}Qwen3\\-32B-T~\cite{yang2025qwen3technicalreport}\end{tabular} 
& \begin{tabular}{@{}c@{}}\taxonomymodellogo{seed}UI-TARS\\-1.5-7B~\cite{qin2025ui}\end{tabular} 
& \begin{tabular}{@{}c@{}}\taxonomymodellogo{claude}Claude\\-Opus-4.6~\cite{anthropic2026claude46}\end{tabular} 
& \begin{tabular}{@{}c@{}}\taxonomymodellogo{gemini}Gemini3\\-flash~\cite{gemini3_2026}\end{tabular} 
& \begin{tabular}{@{}c@{}}\taxonomymodellogo{meituan}EvoCUA\\-32B~\cite{xue2026evocuaevolvingcomputeruse}\end{tabular} \\
\midrule

\multicolumn{11}{c}{\textbf{Procedural Setup and File Management}} \\
\midrule
\rowcolor{rowgreen}
Generative Workflow (GW)
& 1.000 & 1.000 & 1.000 & 1.000 & 1.000 
& 1.000 & 1.000 & 1.000 & 1.000 & 1.000 \\

\rowcolor{rowgreen}
Export and Delivery (ED)
& 0.750 & 0.917 & \textbf{1.000} & 0.917 & 0.917 
& 0.767 & 0.833 & 0.967 & \textbf{1.000} & 0.900 \\

\rowcolor{rowgreen}
Launch and Setup (LS)
& 0.900 & 0.900 & 0.933 & \textbf{0.967} & 0.767 
& 0.803 & 0.752 & \textbf{0.897} & 0.872 & 0.786 \\

\rowcolor{rowgreen}
Preview, Check, and Validation (PCV)
& \textbf{0.800} & 0.600 & \textbf{0.800} & \textbf{0.800} & \textbf{0.800} 
& 0.737 & 0.579 & \textbf{0.947} & 0.842 & 0.737 \\

\rowcolor{rowgreen}
Asset Import and Management (AIM)
& 0.421 & 0.333 & \textbf{0.719} & 0.667 & 0.456 
& 0.605 & 0.542 & \textbf{0.814} & 0.757 & 0.588 \\

\rowcolor{rowgreen}
\textbf{Average (Procedural)}
& 0.774 & 0.750 & \textbf{0.890} & 0.870 & 0.788 
& 0.782 & 0.741 & 0.925 & \textbf{0.894} & 0.802 \\

\midrule
\multicolumn{11}{c}{\textbf{Core Media Editing and Processing}} \\
\midrule

\rowcolor{roworange}
Timeline Editing and Arrangement (TEA) 
& 0.550 & 0.350 & 0.600 & \textbf{0.650} & 0.550 
& 0.359 & 0.333 & \textbf{0.577} & 0.538 & 0.295 \\

\rowcolor{roworange}
Effects and Visual Tuning (EVT)
& 0.207 & 0.276 & \textbf{0.586} & 0.483 & 0.310 
& 0.232 & 0.183 & \textbf{0.537} & 0.488 & 0.415 \\

\rowcolor{roworange}
Audio and Rhythm Editing (ARE)
& 0.167 & 0.167 & 0.333 & \textbf{0.500} & 0.333 
& 0.643 & 0.429 & \textbf{0.929} & 0.786 & 0.643 \\

\rowcolor{roworange}
Masking, Matting, and Tracking (MMT)
& 0.143 & 0.095 & 0.286 & \textbf{0.381} & 0.238 
& 0.368 & 0.439 & \textbf{0.649} & 0.605 & 0.395 \\

  \rowcolor{roworange}
  \textbf{Average (Core Editing)}
  & 0.267 & 0.222 & 0.451 & \textbf{0.504} & 0.358 
  & 0.400 & 0.346 & \textbf{0.673} & 0.604 & 0.437 \\
\midrule
\multicolumn{11}{c}{\textbf{Overall Performance}} \\
\midrule

\rowcolor{rowblue}
\textbf{Average (Overall)} 
& 0.484 & 0.441 & \textbf{0.683} & 0.672 & 0.516 
& 0.532 & 0.502 & \textbf{0.748} & 0.704 & 0.552 \\

\bottomrule
\end{tabular}
}
\vspace{-4mm}
\end{table*}

Given the open-ended and temporally extended nature of professional media post-production, CutVerse mandates \textit{online execution} as its foundational evaluation paradigm. As \figref{task_complexity} illustrates, core editing workflows exhibit substantially longer execution horizons and higher interaction density than procedural tasks, rendering static action prediction fundamentally insufficient. Instead, agents are deployed within live, resettable Windows virtual machines. In this closed-loop environment, agents continuously perceive high-density 
\begin{wrapfigure}{r}{0.50\textwidth}
  \vspace{-3mm} 
  \centering
  \includegraphics[width=0.95\linewidth]{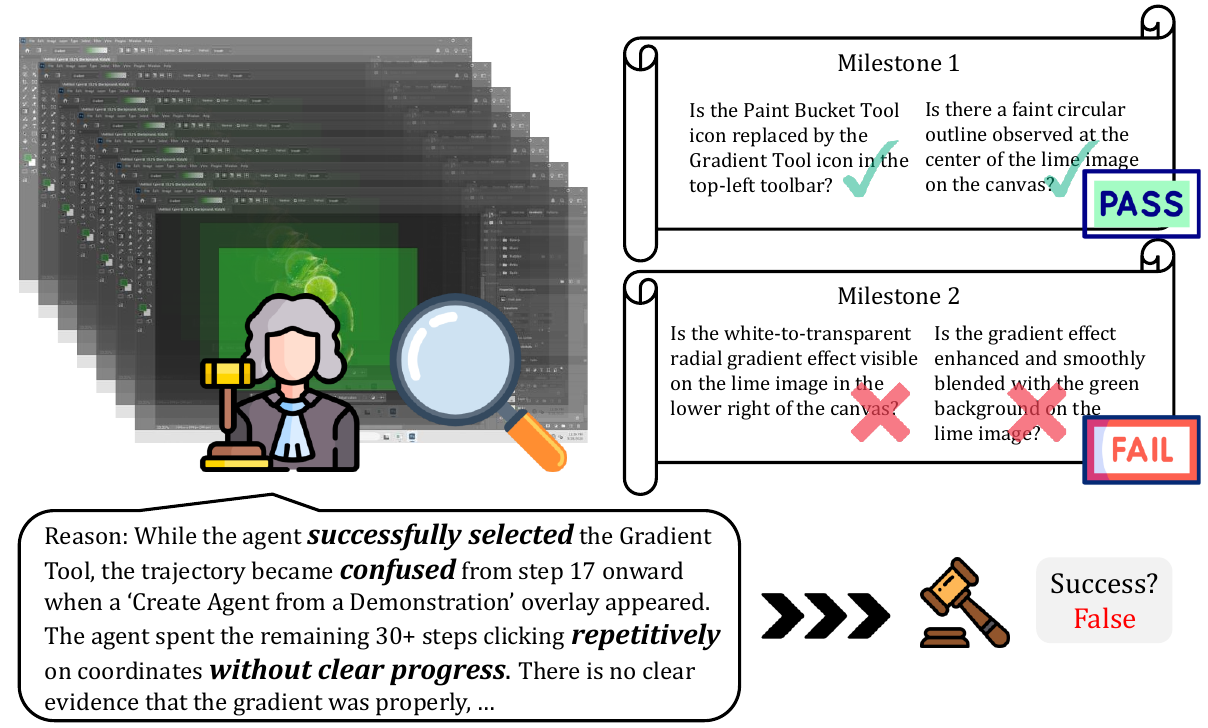}
  \caption{\textbf{Core media editing tasks require longer horizons.} Left: average duration. Right: average steps. Orange bars are core media editing and processing tasks, which are generally higher than procedural tasks in both duration and step count.}
  \label{fig:task_complexity}
  \vspace{-3mm} 
\end{wrapfigure}
workspaces—coupling dynamic video canvases, multi-track timelines, and audio waveforms—to iteratively issue low-level actions. This rigorously tests an agent's capacity to resolve visual ambiguity, maintain temporal synchronization, and autonomously recover from cascading errors over prolonged trajectories.

While imperative for authentic benchmarking, online execution introduces a formidable bottleneck: evaluating open-ended, multimodal outcomes lacking deterministic programmatic verification. Unlike software engineering benchmarks governed by unit tests, operations in creative suites — such as continuous timeline dragging or non-linear spatial adjustments—produce highly contextual, non-symbolic outcomes that completely defy hard-coded heuristics. To alleviate this, we formulate a \textit{Milestone-driven Automated Evaluation Protocol}. We decompose monolithic task trajectories into a hierarchical sequence of semantically meaningful milestones, each encapsulating a verifiable audio-visual state transition. We then orchestrate a scalable VLM-as-a-Judge pipeline, evaluating agent progress via grounded question-answer (QA) pairs aligned with intermediate editing states. This transforms ill-posed trajectory comparison into interpretable, fine-grained verification steps.

To mitigate evaluator hallucination and architectural bias, we instantiate this protocol across distinct frontier vision-language models (i.e., GPT-5.4~\cite{openai_gpt5} and Claude-4.6-Opus~\cite{anthropic2026claude46}). This multi-model grounding ensures that milestone verification relies on robust, model-agnostic visual comprehension rather than evaluator-specific leniency, establishing a principled foundation for assessing complex audio-visual transformations.

To empirically validate this automated infrastructure, we conducted a comprehensive human-alignment study across 300 agent-executed trajectories. Parallel assessments by professional creators and our QA-grounded VLM evaluators demonstrated exceptional concordance: a 98.3\% human-agreement rate using GPT-5.4~\cite{openai_gpt5}, and 99\% with Claude-4.6-Opus~\cite{anthropic2026claude46}. These findings unequivocally prove that our milestone protocol empowers automated models to match expert-level judgment. Consequently, CutVerse successfully relegates human experts to the role of ultimate ground-truth curators, establishing a fully scalable, reproducible, and scientifically rigorous evaluation pipeline for multimodal agentic systems.

\section{Baseline}
\label{sec:Baseline}

\subsection{Baselines Setup}

We benchmark a diverse set of current large-scale vision-language models in CutVerse under a unified online execution framework. Specifically, we evaluate state-of-the-art proprietary models, namely Claude-Opus-4.6~\cite{anthropic2026claude46} and Gemini-3-flash~\cite{gemini3_2026}, accessed via their official APIs. Concurrently, we locally deploy leading open-source models, including Qwen3-32B~\cite{yang2025qwen3technicalreport}, UI-TARS-1.5-7B~\cite{qin2025ui}, and EvoCUA-32B~\cite{xue2026evocuaevolvingcomputeruse}, on a hardware cluster equipped with four NVIDIA RTX 5090 GPUs. All models are prompted to generate structured GUI actions and tool calls based on task descriptions and visual observations. Evaluations are conducted exclusively in an online setting, where agents interact with a live environment and perceive real-time screenshots and state changes. To ensure rigorous and fair comparisons, all models execute the identical set of tasks within completely standardized Windows 11 Pro virtual machines powered by Hyper-V. Each testing episode is strictly initialized with the exact same system states, source files, input formats, and software configurations.

\Paragraph{Evaluation Setting.}
We evaluate under a task-level action execution setting that reflects realistic agent deployment. At each step, the model receives the overall high-level task instruction, the historical context consisting of the last k=$5$ actual execution screenshots alongside their natural language descriptions and pyautogui code, and the current keyframe screenshot. Crucially, we move beyond passive prediction. Our framework requires the agent to actually execute the inferred pyautogui operations directly within the live virtual machine. The model must autonomously determine the next action based solely on the task goal and multimodal history without relying on step-level instructions. This closed-loop setup accurately mirrors how autonomous agents operate within practical post-production workflows.

\begin{table}[!t]
\centering
\caption{
\captfont{
Task execution accuracy by software across models, augmented with benchmark complexity statistics.
}
}
\vspace{-3mm}
\label{tab:software_all}

\footnotesize 

\setlength{\tabcolsep}{6pt} 
\renewcommand{\arraystretch}{1.2}

\begin{tabular}{lccccccc}
\toprule
\rowcolor{headerblue}
\textbf{Software} & \textbf{AvgSteps} & \textbf{AvgDur(s)} & \taxonomymodellogo{Claude}\textbf{Claude} & \taxonomymodellogo{gemini}\textbf{Gemini} & \taxonomymodellogo{meituan}\textbf{Evo} & \taxonomymodellogo{Qwen}\textbf{Qwen} & \taxonomymodellogo{seed}\textbf{UI-TARS}\\
\midrule
\taxonomymodellogo{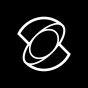}Keling & 8.31 & 26.56 & 0.815 & 0.852 & 0.704 & 0.852 & 0.593 \\
\taxonomymodellogo{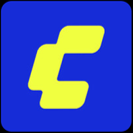}ComfyUI & 10.33 & 33.30 & 0.667 & 0.833 & 0.500 & 0.500 & 0.500 \\
\taxonomymodellogo{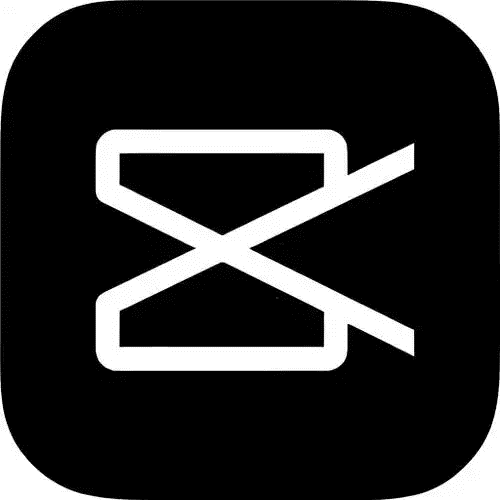}JianYing & 22.33 & 63.32 & 0.754 & 0.725 & 0.493 & 0.522 & 0.493 \\
\taxonomymodellogo{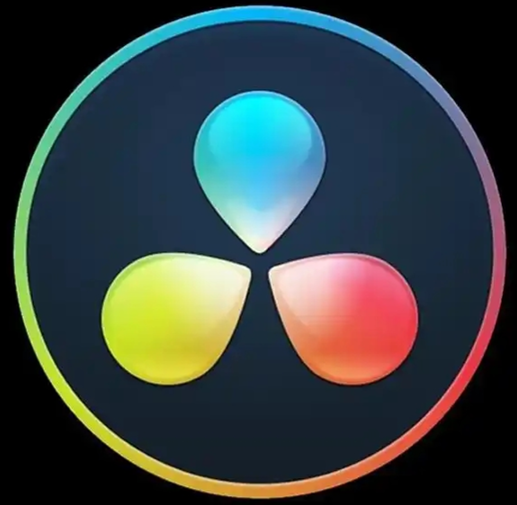}DaVinci & 16.60 & 46.68 & 0.750 & 0.700 & 0.550 & 0.450 & 0.450 \\
\taxonomymodellogo{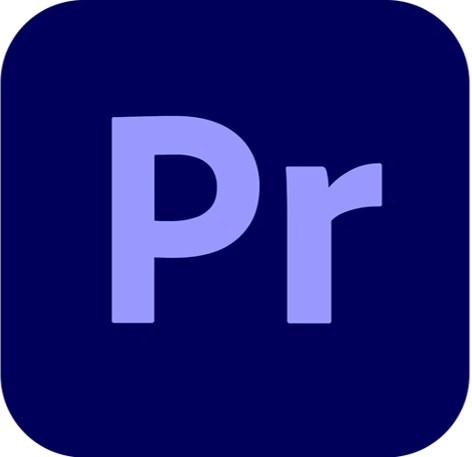}Premiere Pro & 12.98 & 26.35 & 0.642 & 0.660 & 0.604 & 0.491 & 0.396 \\
\taxonomymodellogo{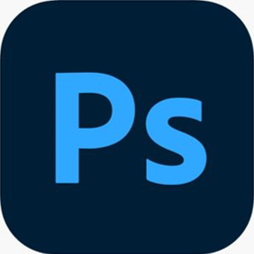}Photoshop & 42.61 & 91.20 & 0.576 & 0.576 & 0.455 & 0.424 & 0.455 \\
\taxonomymodellogo{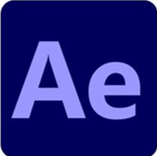}After Effects & 14.81 & 47.44 & 0.577 & 0.500 & 0.269 & 0.269 & 0.346 \\
\bottomrule
\end{tabular}

\vspace{-6mm} 
\end{table}

\subsection{Results}
\Paragraph{High Proficiency in Procedural Operations.}
As presented in \tabref{unified_analysis_task_milestone}, all evaluated models exhibit robust capabilities within the procedural setup and file management category. Notably, every model achieves a perfect success rate of 1.000 at both the task and milestone levels for generative workflows. Performance remains highly competitive in operations such as export and delivery, as well as launch and setup. In these areas, Claude-Opus-4.6~\cite{anthropic2026claude46} and Gemini-3-flash~\cite{gemini3_2026} consistently outpace the other models. Although performance slightly decreases in asset import and management, the overall success rates in this upper section of the table indicate a strong baseline for basic software navigation.

\Paragraph{Severe Degradation in Core Media Editing.}
Conversely, the data reveals a drastic performance drop across all models when transitioning to core media editing and processing tasks. The task success rates plummet in domains requiring precise content manipulation. For instance, in masking, matting, and tracking tasks, the task success rate drops to 0.095 for UI-TARS~\cite{qin2025ui}. Even the top-performing models struggle significantly in this category, with Claude-Opus-4.6~\cite{anthropic2026claude46} and Gemini-3-flash~\cite{gemini3_2026} scoring only 0.286 and 0.381, respectively. Similarly, effects and visual tuning tasks yield extremely low task success rates, bottoming out at 0.207 for Qwen3-32B~\cite{yang2025qwen3technicalreport}, which clearly illustrates the complexity of these operations.

\Paragraph{Quantitative Gap Between Milestones and Tasks.}
Furthermore, the dual-metric structure of the table explicitly exposes a substantial numerical gap between local milestone success rates and overall task success rates, particularly within the complex editing categories. A prominent example is observed in the audio and rhythm editing operations. While Claude achieves a remarkably high milestone success rate of 0.929 in this specific category, its overall task success rate falls sharply to 0.333. Gemini similarly drops from a 0.786 milestone success rate to a 0.500 task success rate. This consistent statistical discrepancy across EvoCUA-32B~\cite{xue2026evocuaevolvingcomputeruse}, Qwen3\cite{yang2025qwen3technicalreport}, and UI-TARS-1.5-7B~\cite{qin2025ui} confirms that achieving high accuracy on intermediate procedural steps does not quantitatively translate to the successful completion of the entire multi-step editing task.


\begin{figure}[!t]
    \centering
    \includegraphics[width=1.\linewidth]{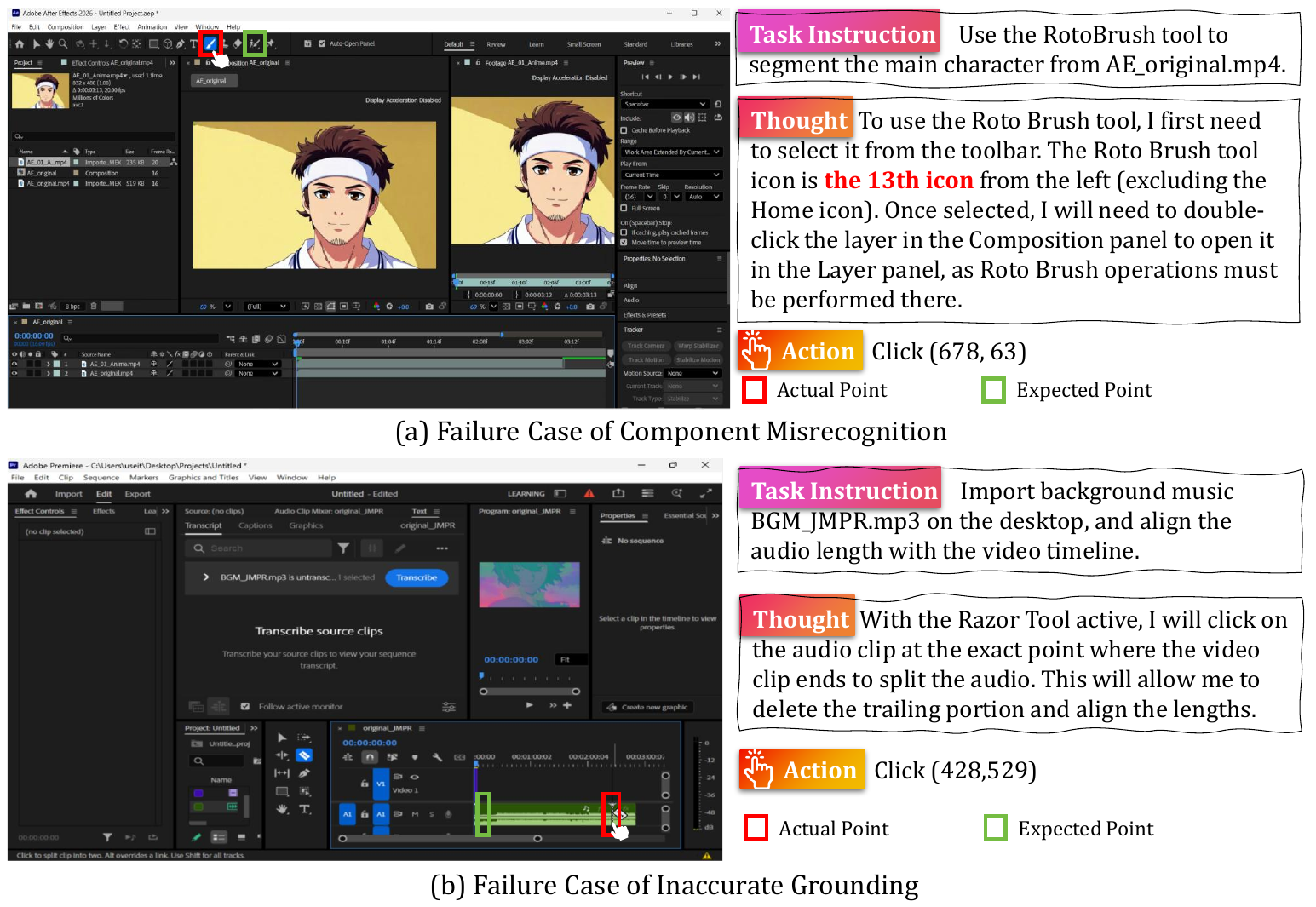} 
\caption{\textbf{Typical visual perception failures.} (a) \textbf{Component Misrecognition}: Agents struggle to identify unlabelled tools in condensed layouts. (b) \textbf{Inaccurate Grounding}: Lack of pixel-level precision prevents delicate timeline operations.}
    \label{fig:FailureCase1}
\end{figure}

\section{Analysis}
\label{sec:Analysis}
In this section, we analyze model behavior in multimodal media settings and identify key bottlenecks of GUI agents.


\begin{figure}[!t]
    \centering
    \begin{minipage}{0.48\textwidth}
        \centering
        \includegraphics[width=\linewidth]{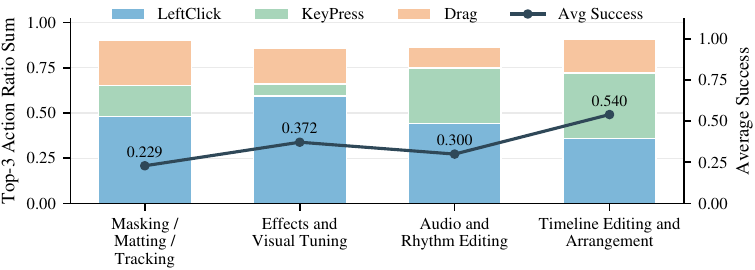} 
        \caption{\textbf{Action distribution vs. success rate in core media editing tasks.} Core editing tasks exhibit a higher reliance on compositional interactions, reflected by the more balanced distributions of top-3 action types.}
        \label{fig:actionspace}
    \end{minipage}\hfill 
    \begin{minipage}{0.48\textwidth}
        \centering
        \includegraphics[width=\linewidth]{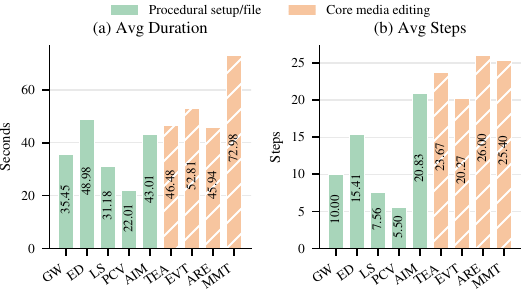} 
        \caption{\textbf{Core media editing tasks require longer horizons.} Left: average duration. Right: average steps. Orange bars are core media editing tasks, which are generally higher than green procedural tasks in both duration and step count.}
        \label{fig:task_complexity}
    \end{minipage}
\end{figure}

\subsection{Milestone-Task Consistency Gap}
As indicated in \tabref{unified_analysis_task_milestone},milestone-level performance consistently exceeds end-to-end task success across all models, indicating that current GUI agents already exhibit basic planning and execution capabilities in media editing workflows. However, this local competence fails to translate into global success. While agents reliably complete procedural milestones such as \textit{Launch}, \textit{Export}, and \textit{Preview}, they struggle on critical stages in core editing workflows, including \textit{Effects and Visual Tuning}, \textit{Audio and Rhythm Editing}, \textit{etc.}

This starkly demonstrates that existing GUI agents fundamentally lack the requisite planning and execution capabilities within the specialized vertical of professional media editing. We present a more comprehensive analysis regarding these domain-specific deficiencies in the subsequent sections.

\begin{figure*}[!t]
    \centering
    \includegraphics[width=1.\textwidth]{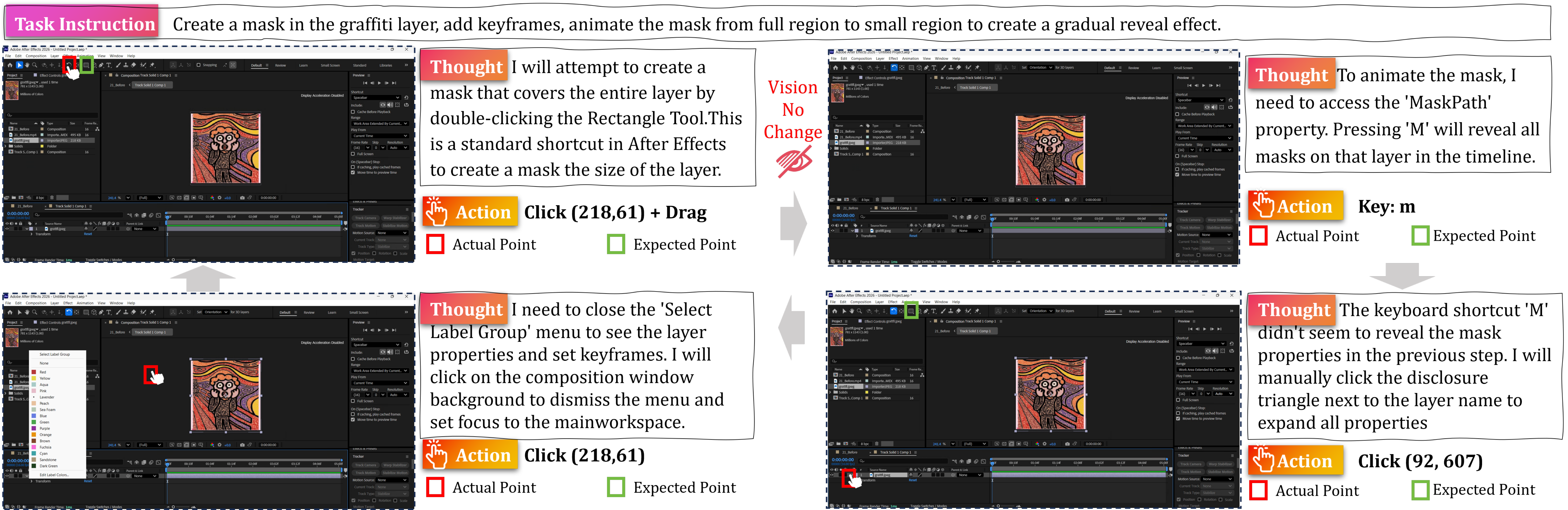}
    \caption{
\textbf{Infinite loops from static feedback.} Lacking obvious visual alterations ("Vision No Change"), the agent fails to register state transitions. This perceptual blind spot traps the model in repetitive cycles of redundant clicks, halting progress. Red/green boxes denote actual/expected targets.
    }
    \label{fig:Qualitity_c}
\end{figure*}

\subsection{Media Applications Complexity}
As shown in \tabref{software_all}, models achieve high accuracy on structured generation tools (\eg ComfyUI) but degrade substantially on professional editing software (\eg After Effects). These professional environments impose dense visual layouts and demand sustained, long-horizon interactive operations.
This disparity exposes a critical bottleneck: current agents fail to maintain robust \textit{cross-modal alignment} and \textit{audio-visual grounding} under multimodal information overload. Consequently, software complexity serves as a direct proxy for multimodal reasoning difficulty.

\subsection{Long-Horizon Multimodal Task Difficulty}
As illustrated in \figref{task_complexity}, task structural complexity directly dictates agent performance. Core media editing operations demand significantly longer execution horizons and higher step densities than procedural tasks. For instance, masking and tracking operations average nearly 73 seconds across 25 atomic steps, whereas procedural previewing requires merely 22 seconds and fewer than 6 steps.

This extended temporal horizon fundamentally exacerbates multimodal complexity. Throughout lengthy sequences, agents must continuously align evolving visual layouts, audio signals, and latent editing intents across dozens of interface states. Maintaining this cross-modal consistency proves highly challenging, as minor perceptual or planning errors compound irreversibly over time. This error accumulation ultimately drives the systemic failures and high incomplete ratios inherent to core media editing tasks.

\subsection{Missing Compositional Action Space}
As shown in \figref{actionspace}, core media editing tasks exhibit more balanced and diverse action distributions across \textit{LeftClick}, \textit{KeyPress}, and \textit{Drag}, yet their success rates remain consistently low. 
This mismatch indicates that the challenge lies not in action availability, but in action coordination. 
Such tasks inherently require tightly coupled, compositional interactions (\eg key–mouse combinations and temporally synchronized operations), which cannot be decomposed into independent atomic steps. 
As a result, current action space of GUI Agent fundamentally limit executability in complex editing workflows.

\begin{figure}[!t]
\centering
\includegraphics[width=1.\linewidth]{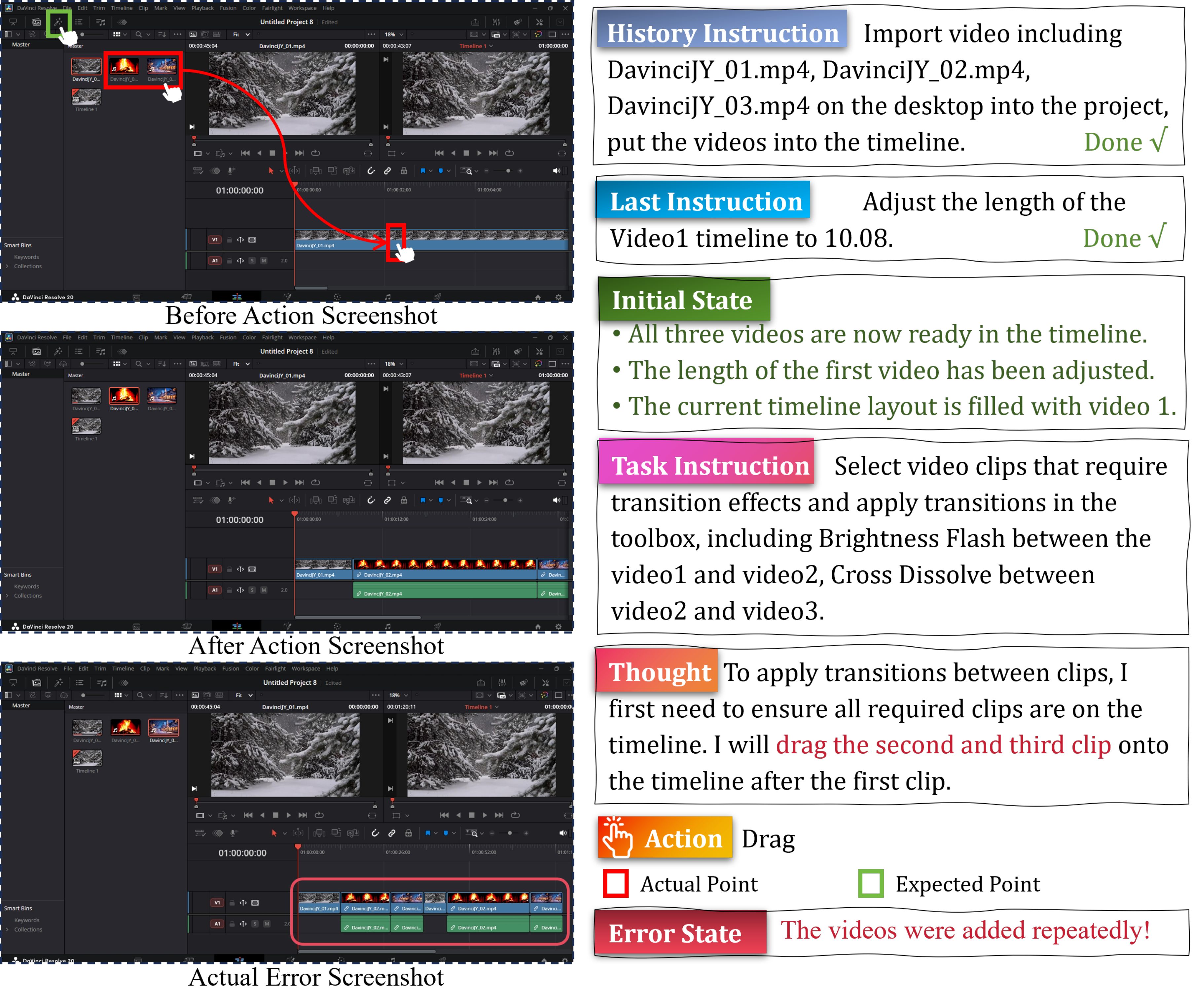}

\caption{\textbf{Global state neglect.} Confined to a localized, zoomed-in timeline view, the agent misses the macro-level context. This myopic perception falsely suggests missing clips, triggering redundant drag operations that erroneously duplicate existing assets.}
\label{fig:Qualitive_d}

\end{figure}

\subsection{Qualitative Evaluation}
To elucidate the specific failure modes of current vision-language models, we conduct a qualitative analysis of their interactive execution trajectories which reveals four critical behavioral deficiencies that restrict their deployment in professional editing workflows.

\Paragraph{Component Misrecognition and Blind Spots.}
Professional media software features highly condensed interfaces populated with a massive array of specialized functional components. However, evaluated agents predominantly recognize universally common icons or buttons accompanied by explicit text labels.As illustrated in \figref{FailureCase1} (a) They frequently fail to identify domain-specific tools, unlabelled toolbars, or subtle interface elements, severely limiting their ability to utilize advanced software features.

\Paragraph{Inaccurate Fine-Grained Grounding.}
Current graphical user interface agents struggle significantly with precise spatial localization. As illustrated in \figref{FailureCase1} (b), when tasks require pinpoint accuracy on a video timeline or exact coordinate selection for specific canvas elements, models frequently miss the intended target. This lack of pixel-level visual grounding prevents agents from performing delicate temporal trimmings or precise spatial adjustments.

\Paragraph{Lack of Global Perception.} 
Current agents lack proactive visual exploration, relying on localized observations rather than verifying the global workspace state. 
This behavior is intrinsically linked to the \textit{Missing Compositional Action Space}. Because operations like global zooming require complex key-mouse coordination, agents are mechanically restricted from acquiring macro-level context, inevitably triggering erroneous operations based on incomplete information.

\Paragraph{Repetitive Action Loops Triggered by Static Visual Feedback.}
Agents rely heavily on immediate visual confirmation to verify state transitions. When an executed action produces no obvious visual alteration in the subsequent interface screenshot, the model fails to register the system state change. Consequently, the agent repeatedly issues the exact same historical action commands, ultimately trapping the execution process in an infinite operational loop, as shown in \figref{Qualitity_c}.

\section{Conclusion}
\label{sec:conclusion}

This work presents \textbf{CutVerse}, the first systematic benchmark and scalable infrastructure for systematically evaluating computer-use agents within real-world media post-production workflows. Our findings expose a profound gap between current agent capabilities and professional creative demands. While agents handle structured procedural operations, they systematically fail during the sustained execution of complex editing tasks that necessitate precise spatial grounding, temporal coordination, and compositional control. These limitations strictly underscore the necessity of authentic, high-fidelity evaluation environments. Ultimately, it is our vision that CutVerse will serve as a practical foundation for advancing end-to-end multimedia production.

\addtocontents{toc}{\protect\setcounter{tocdepth}{1}}
\clearpage
\beginappendix
\lstdefinestyle{prompttemplate}{
    basicstyle=\ttfamily\scriptsize\linespread{0.8}\selectfont,
    breaklines=true,
    breakatwhitespace=false,
    breakautoindent=false,
    breakindent=0pt,
    columns=fullflexible,
    keepspaces=true,
    frame=single
}

\lstdefinestyle{gamepromptblock}{
    basicstyle=\ttfamily\scriptsize\linespread{0.8}\selectfont,
    breaklines=true,
    breakatwhitespace=false,
    columns=fullflexible,
    keepspaces=true,
    frame=single,
    framesep=3pt,
    aboveskip=0.25em,
    belowskip=0.25em
}

\definecolor{jsonkey}{HTML}{2B59C3}
\definecolor{jsonstring}{HTML}{0B7A75}
\definecolor{jsonnumber}{HTML}{D94A7A}
\definecolor{jsonkeyword}{HTML}{7C4DFF}
\definecolor{jsondelim}{HTML}{7A3E00}
\definecolor{jsonpunct}{HTML}{7A7A7A}

\lstdefinelanguage{json}{
    basicstyle=\ttfamily\scriptsize\linespread{0.8}\selectfont,
    showstringspaces=false,
    breaklines=true,
    breakatwhitespace=false,
    columns=fullflexible,
    keepspaces=true,
    upquote=true,
    morestring=[b]",
    stringstyle=\color{jsonstring},
    keywordstyle=\color{jsonkeyword}\bfseries,
    keywords={true,false,null},
    moredelim=[s][\color{jsonkey}]{"}{":},
    literate=
     *{0}{{{\color{jsonnumber}0}}}{1}
      {1}{{{\color{jsonnumber}1}}}{1}
      {2}{{{\color{jsonnumber}2}}}{1}
      {3}{{{\color{jsonnumber}3}}}{1}
      {4}{{{\color{jsonnumber}4}}}{1}
      {5}{{{\color{jsonnumber}5}}}{1}
      {6}{{{\color{jsonnumber}6}}}{1}
      {7}{{{\color{jsonnumber}7}}}{1}
      {8}{{{\color{jsonnumber}8}}}{1}
      {9}{{{\color{jsonnumber}9}}}{1}
      {.}{{{\color{jsonnumber}{.}}}}{1}
      {-}{{{\color{jsonnumber}{-}}}}{1}
      {:}{{{\color{jsonpunct}{:}}}}{1}
      {,}{{{\color{jsonpunct}{,}}}}{1}
      {\{}{{{\color{jsondelim}{\{}}}}{1}
      {\}}{{{\color{jsondelim}{\}}}}}{1}
      {[}{{{\color{jsondelim}{[}}}}{1}
      {]}{{{\color{jsondelim}{]}}}}{1},
}

\lstdefinestyle{jsonblock}{
    language=json,
    basicstyle=\ttfamily\scriptsize\linespread{1}\selectfont,
    showstringspaces=false,
    breaklines=true,
    breakatwhitespace=false,
    columns=fullflexible,
    keepspaces=true,
    frame=single,
    framesep=3pt,
    rulecolor=\color{headerblue},
    backgroundcolor=\color{bggray}
}

\section{Details for Benchmark}
\label{sec:benchmark_details}

\begin{table}[htbp]
\centering
\small
\caption{Unified atomic action space combining standard operating system actions and real user interaction traces. The vocabulary is strictly limited to low-level mouse and keyboard operations to enforce pure visual grounding.}
\label{tab:action_space}
\setlength{\tabcolsep}{4pt}
\renewcommand{\arraystretch}{1.15}
\begin{tabular}{l p{0.68\linewidth}}
\toprule
\textbf{Function} & \textbf{Description} \\
\midrule
\texttt{moveTo($x, y$)} & Moves the mouse cursor to the specified screen coordinates. \\
\texttt{click($x, y$)} & Performs a left mouse click at the given coordinates. \\
\texttt{dragTo($x, y$)} & Drags the mouse cursor to the target position while holding the mouse button. \\
\texttt{scroll($\Delta$)} & Scrolls the interface vertically by a given offset. \\
\texttt{write(text)} & Types the specified text at the current cursor location. \\
\texttt{keyDown($k$)} & Presses and holds a keyboard key (such as \texttt{Ctrl}). \\
\texttt{keyUp($k$)} & Releases a previously pressed keyboard key. \\
\texttt{keyPress($k$)} & Presses and releases a keyboard key as a single atomic action. \\
\texttt{hotkey($k_1, k_2$)} & Executes a keyboard shortcut combination such as \texttt{Ctrl+C}. \\
\texttt{WAIT} & Agent pauses execution and waits for observable environment changes. \\
\texttt{DONE} & Agent declares that the task has been successfully completed. \\
\texttt{FAIL} & Agent determines that the task is infeasible or cannot be completed. \\
\bottomrule
\end{tabular}
\end{table}

To ensure complete transparency and facilitate future research, this section provides a comprehensive breakdown of the \textbf{CutVerse} benchmark. We detail the rigorous human annotation protocol, formally define our multimodal compositional action space, and present the granular specifications for the 186 evaluation tasks across the 7 professional software platforms.

\subsection{Human Annotation Protocol}
\label{subsec:annotation}

Ensuring high-fidelity expert trajectories is critical for rigorously evaluating computer-use agents in professional workflows. To construct our dataset, we recruited a dedicated cohort of 10 professional creators possessing extensive expertise in both traditional post-production software and AIGC workflows. These experts meticulously authored the foundational data for all 186 tasks, a comprehensive procedure encompassing formal task definition, the recording of ground-truth execution videos guided by authentic task instructions, and the instantiation of dedicated Virtual Machine (VM) checkpoints for environment standardization. Following the raw data collection, the recorded execution videos were systematically processed through our proposed multimodal parsing pipeline infrastructure. This pipeline autonomously parsed the continuous interaction traces to extract high-level task milestones alongside the granular operational content of each individual step. Furthermore, we leveraged Large Language Models (LLMs) in conjunction with pre-action and post-action screenshots to generate contextually rich multimodal Question-Answer (QA) pairs. Finally, to guarantee utmost data integrity, the pipeline concluded with a rigorous human-in-the-loop refinement phase: the original expert recorders manually evaluated the quality of the extracted milestones and generated QA pairs, iteratively adjusting the textual details to ensure absolute semantic precision and alignment with the visual trajectories.
Each standalone benchmark run is launched using a registry preset passed via the command-line flag \texttt{--config}. The preset has the form:

\subsection{Action Space Definition}
\label{subsec:action_space}

We formulate the continuous GUI interaction as a multimodal Partially Observable Markov Decision Process (POMDP). At each time step $t$, the observation $O_t = (I_t, H_t)$ strictly encapsulates the raw, high-resolution visual interface $I_t \in \mathbb{R}^{H \times W \times 3}$ and the action history $H_t$. By deliberately isolating the agent from structured underlying metadata (e.g., accessibility trees or DOMs), we systematically enforce pure visual grounding. Consequently, this formalization authentically mirrors the inherent complexity of professional multimedia production, compelling the agent to navigate pervasive multimodal information overload exclusively via manual, compositional key-mouse operations without algorithmic shortcuts.

To robustly support this paradigm, the action space $\mathcal{A}$ is meticulously constrained to low-level GUI executions. As detailed in Table~\ref{tab:action_space}, these operations encompass precise pixel-level point-and-click mechanisms (\texttt{click}), sustained spatial controls strictly required for audio-visual temporal synchronization (\texttt{dragTo}), and semantic keyboard inputs for generative cross-modal prompting (\texttt{write}, \texttt{hotkey}). Furthermore, the vocabulary natively integrates macro-level viewport navigation (\texttt{scroll}), a vital capability for agents to actively alleviate spatial blind spots during complex editing. Ultimately, this scalable infrastructure allows us to rigorously benchmark whether an agent can autonomously orchestrate the long-horizon, multimodal coordination mandated by professional creative workflows.

Consequently, the action space $\mathcal{A}$ consists exclusively of low-level GUI operations. As detailed in Table~\ref{tab:action_space}, these actions encompass basic point-and-click operations requiring precise pixel-level coordinates (such as \texttt{click($x, y$)}), continuous spatial controls heavily utilized for timeline adjustments (such as \texttt{dragTo($x, y$)}), and keyboard inputs for shortcuts and parameter tuning (such as \texttt{write(text)} and \texttt{hotkey($k_1, k_2$)}). Furthermore, the vocabulary includes viewport navigation operations like scrolling, which are mandatory for agents to actively explore the macro-level workspace, alongside system control states (\texttt{WAIT}, \texttt{DONE}, \texttt{FAIL}). This precise formulation allows us to evaluate not merely whether an agent triggered a button, but whether it can autonomously sustain the long-horizon key-mouse coordination required by sophisticated professional software.

\subsection{Detailed Task Specifications}
\label{subsec:task_details}

Anchored in authentic multimedia post-production scenarios, our benchmark comprises 186 meticulously curated tasks designed to rigorously evaluate computer-use agents. To ensure comprehensive coverage of real-world creative demands, our expert recorders systematically aggregated a complete taxonomy of fundamental editing typologies, subsequently tailoring specific task instantiations across diverse software environments. These tasks encapsulate the full spectrum of multimodal workflows, ranging from generative cross-modal asset creation to precise audio-visual timeline synchronization and intricate visual tuning. \tabref{task_taxonomy_examples} presents our formalized nine-category taxonomy, illustrating the distinct complexities with concrete, domain-specific examples. The complete, exhaustive list of all 186 tasks, alongside their corresponding initial states and multimodal evaluation criteria, is open-sourced and available in our project repository.

\Paragraph{Milestone-Driven visualization of CutVerse..} 
In professional media post-production, execution workflows are inherently continuous and long-horizon, rendering binary success metrics insufficient for robust evaluation. To address this challenge, we introduce a granular, milestone-driven evaluation framework within CutVerse, as illustrated in Figure~[Insert Figure Number]. Instead of assessing a complex task merely by its final output, our parsing infrastructure systematically decomposes the continuous execution trajectory into discrete, semantic milestones. For instance, the instruction to apply and configure a transition effect is parsed into sequential milestones, such as locating the specific effects panel, dragging the "Cross Dissolve" transition onto the targeted timeline intersection, and precisely adjusting the "Edge Feather" parameter to a specific numerical value. 

Crucially, to automate and rigorously assess the completion of each intermediate step, we formulate a multimodal Question-Answer (QA) verification mechanism. For every defined milestone, context-specific QA pairs are established to interrogate the visual interface before and after the agent's designated action. These QA pairs target precise spatial-temporal state transitions rather than superficial clicks. As demonstrated in the evaluation protocol, the system visually verifies granular interface alterations, such as confirming the appearance of a specialized 'fx' badge on the video clip or validating that the parameter slider in the Effect Controls panel has successfully shifted to the value of 71. By anchoring the evaluation to these visually grounded QA pairs, CutVerse provides an interpretable, step-by-step diagnostic of the agent's multimodal perception and reasoning capabilities, effectively preventing false positives and exposing the exact points of failure within sophisticated creative workflows.
\begin{figure*}[t]
    \centering
    \includegraphics[width=.98\textwidth]{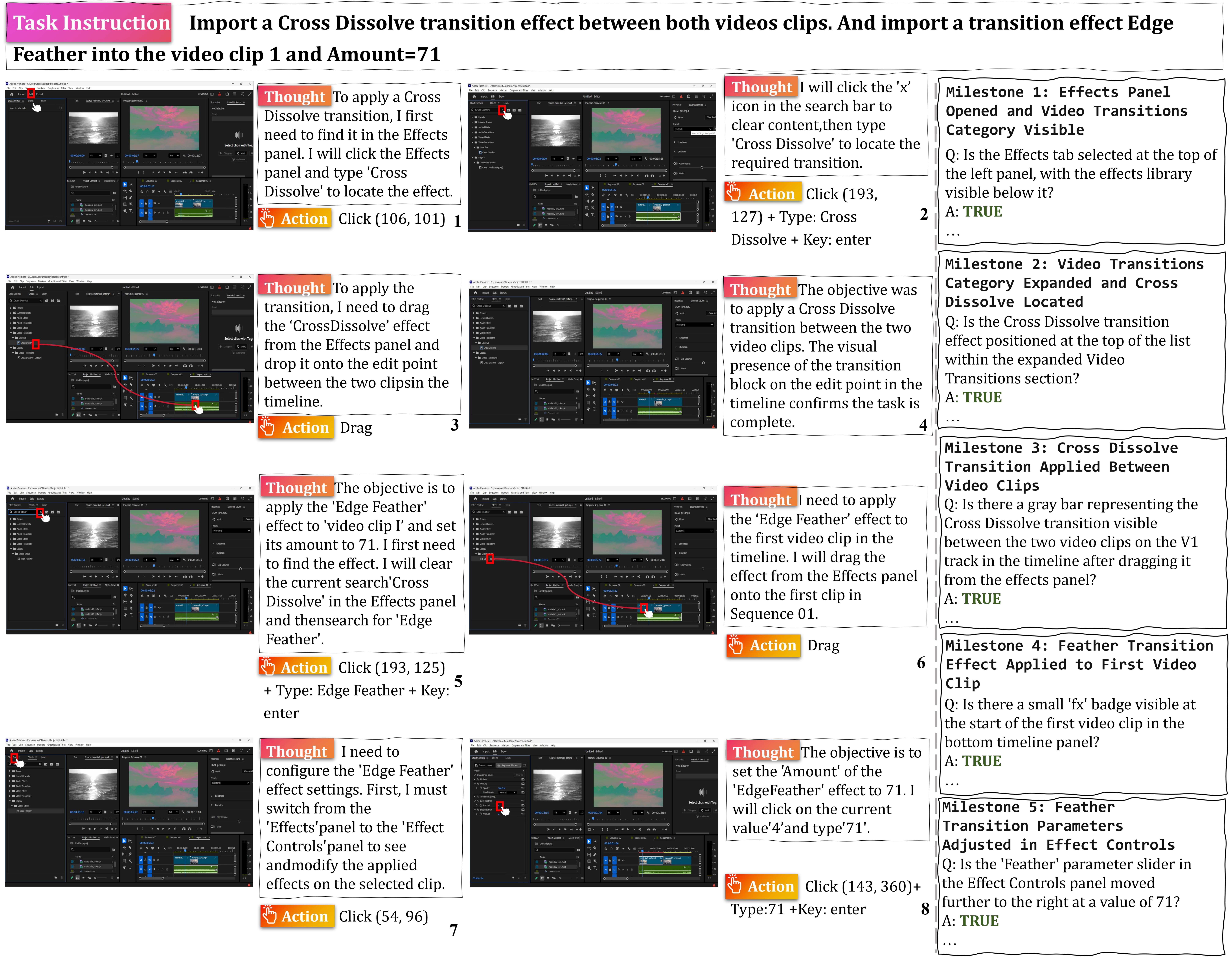}
    \caption{
\textbf{Milestone parsing and multimodal QA verification in CutVerse.} The continuous execution of a complex editing workflow  is systematically decomposed into sequential visual milestones. To rigorously assess agent performance, each milestone is coupled with specific multimodal Question-Answer (QA) pairs. These QA pairs visually interrogate the interface to verify precise spatial-temporal state transitions, ensuring interpretable and highly reliable task evaluation.
    }
    \label{fig:framework}
\end{figure*}

\begin{figure*}[htbp]
    \centering
    \includegraphics[width=.98\textwidth]{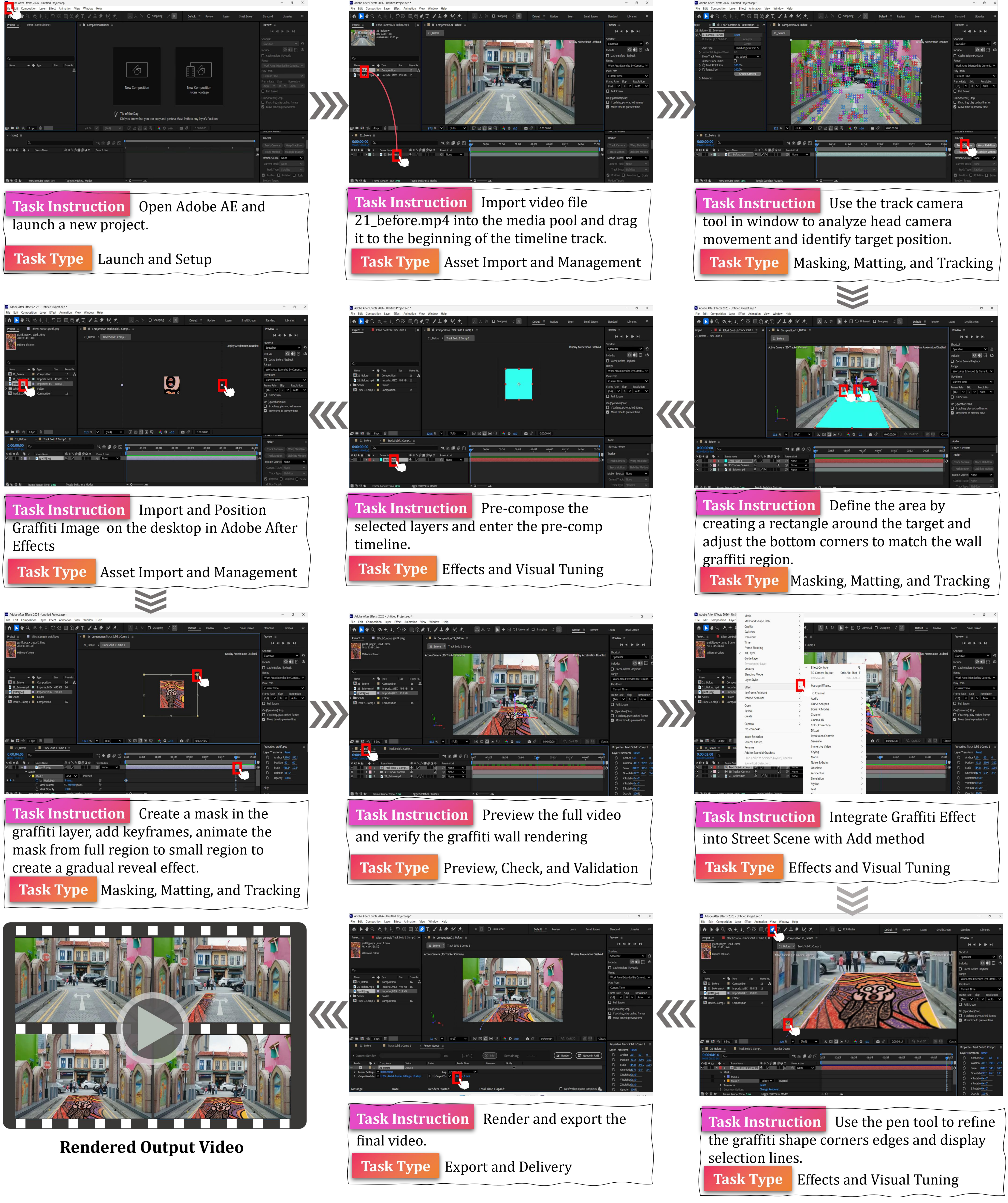}
    \vspace{-3mm}
    \caption{
\textbf{Task-centric decomposition in CutVerse.} Continuous post-production pipelines are systematically deconstructed into standardized, discrete tasks. Mapping high-level creative intents to quantifiable actions enables the granular evaluation of an agent's compositional execution capabilities.
    }
    \label{fig:framework}
\end{figure*}

\Paragraph{Task-Centric visualization of CutVerse.} 
To rigorously evaluate the sustained operational capabilities of computer-use agents, CutVerse systematically deconstructs long-horizon media production pipelines into discrete, categorical tasks. As illustrated in Figure~[Insert Figure Number], a complex creative objective—such as integrating a dynamically tracked asset into a moving sequence—is not evaluated as a monolithic black box. Instead, the continuous execution is meticulously mapped onto our formalized taxonomy of fundamental editing typologies. This approach isolates specific multimodal challenges, tracing the agent's progression from initial \textit{Asset import and management} to advanced \textit{Masking, matting, and tracking}, and culminating in \textit{Export and delivery}. By isolating these atomic components, our framework can pinpoint precisely where an agent's spatial reasoning, temporal synchronization, or functional understanding fails during a sustained editing session. Furthermore, this task-level granularity ensures that the benchmark goes beyond assessing rote memorization of procedural interface clicks; it rigorously evaluates the agent's capacity to autonomously orchestrate diverse, cross-modal operations. Ultimately, this methodology provides a high-resolution diagnostic instrument, exposing the strict boundaries of current multimodal foundation models when confronted with the compositional complexity inherent to professional software environments.

\begin{table*}[!htbp]
\centering
\caption{Summary of agent configurations in the CutVerse benchmark.}
\label{tab:agent_config}
\resizebox{\textwidth}{!}{%
\begin{tabular}{l c c c c c c c c}
\toprule
\textbf{Agent} & \textbf{Size} & \textbf{Backend} & \textbf{Prompt} & \textbf{Image Size} & \textbf{Coord.} & \textbf{Output} & \textbf{History} & \textbf{Think} \\
\midrule
Claude Opus 4.6~\cite{anthropic2026claude46}  & Closed & API     & OSWorld  & $1280\times720$           & Abs.\ (px)          & JSON           & 10 imgs  & --          \\
Qwen3-VL-32B-T\cite{yang2025qwen3technicalreport}    & 32B    & vLLM    & OSWorld  & Smart-resize              & Rel.\ (0--999)      & XML Tool Call  & 4 turns  & \checkmark  \\
UITars 1.5\cite{qin2025ui}        & --     & vLLM    & OSWorld  & $1920\times1080$          & Norm.\ (0--1000)    & Python-style   & 5 imgs   & \checkmark  \\
EvoCUA\cite{xue2026evocuaevolvingcomputeruse}            & 32B    & vLLM    & OSWorld  & Smart-resize              & Rel.\ (0--999)      & CoT+Code / XML & 4 turns  & --          \\
Gemini 3 Flash\cite{gemini3_2026}    & Closed & API     & CutVerse & $\leq$1280 (long side)    & Norm.\ (0--1000)    & JSON           & 5 imgs   & Optional    \\
\bottomrule
\end{tabular}%
}
\end{table*}
\subsection{Agent Implementation Details}
\label{sec:supp_agent_impl}

To ensure a rigorous and reproducible evaluation of GUI agents across the CutVerse benchmark, we integrate five state-of-the-art multimodal foundation models, each instantiated as an autonomous agent with a carefully designed system prompt that governs its perception--reasoning--action loop. All agents share a unified execution pipeline: at each step, the agent receives a screenshot of the current desktop state (resized to a model-specific resolution), along with the task instruction and, where applicable, a sliding window of multi-turn interaction history. The agent then produces a structured action prediction that is parsed and executed by our centralized action engine.

\Paragraph{Prompt Design Rationale.}
For four of the five agents---Claude Opus 4.6~\cite{anthropic2026claude46}\footnote{\url{https://www.anthropic.com/claude/opus}}, Qwen3-VL~\cite{yang2025qwen3technicalreport}\footnote{\url{https://huggingface.co/Qwen/Qwen3-VL-32B-Thinking}}, UITars~1.5~\cite{qin2025ui}\footnote{\url{https://huggingface.co/ByteDance-Seed/UI-TARS-1.5-7B}}, and EvoCUA~\cite{xue2026evocuaevolvingcomputeruse}\footnote{\url{https://huggingface.co/meituan/EvoCUA-32B-20260105}}---we adopt the prompt paradigm established by the OSWorld benchmark~\cite{xie2024osworld}\footnote{\url{https://github.com/xlang-ai/OSWorld}}, which has become the \emph{de facto} standard for desktop GUI agent evaluation. The OSWorld-style prompt defines a \texttt{computer\_use} tool schema that enumerates a canonical action vocabulary (click, type, scroll, key, drag, wait, terminate, \emph{etc.}), specifies a structured output format (either XML tool-call blocks, JSON objects, or PyAutoGUI code), and provides descriptive guidelines for coordinate usage, screenshot consultation, and cursor alignment. We preserve these conventions to ensure a fair, apples-to-apples comparison: every agent is evaluated under the same prompt contract that governs perception, reasoning, and action emission, differing only in model-specific adaptations (e.g., coordinate resolution, image preprocessing, and output parser). In contrast, the Gemini~3~Flash~\cite{gemini3_2026}\footnote{\url{https://aistudio.google.com/models/gemini-3}} agent employs a distinct \emph{unified planner JSON schema} developed specifically for the CutVerse autonomous pipeline, which introduces a normalised 0--1000 coordinate space, compound multi-action arrays, and milestone-completion metadata not present in the OSWorld template.

Below, we provide a comprehensive account of each agent's prompt design, coordinate convention, and output schema, enabling full reproducibility of our benchmark results.

\subsubsection{Claude Opus 4.6 (Anthropic)}
\label{sec:supp_claude}

We employ Claude Opus 4.6 through the native Anthropic Messages API (proprietary, closed-weight). Following the OSWorld prompt paradigm~\cite{xie2024osworld}, we define the standard \texttt{computer\_use} action vocabulary but adapt the output format to a \emph{JSON-only mode}: the system prompt explicitly instructs the model to respond with a single, valid JSON object containing structured fields for observation, chain-of-thought reasoning, action type, coordinates, and a milestone-completion flag. This design eliminates the need for fragile free-text parsing and enforces a deterministic output schema that is directly consumable by the downstream action executor.

\Paragraph{Coordinate Convention.}
Screenshots are resized to a fixed $1280\times720$ pixel canvas before being sent to the model. Coordinates in the JSON response are specified in this pixel space and subsequently rescaled to the actual screen resolution via linear interpolation: $x_{\text{screen}} = \lfloor x \cdot w_{\text{screen}} / 1280 \rfloor$, $y_{\text{screen}} = \lfloor y \cdot h_{\text{screen}} / 720 \rfloor$.

\Paragraph{Multi-turn History.}
Up to 10 most recent screenshots are retained in the conversation context; older turns are replaced with a textual placeholder (\texttt{[previous screenshot]}) to manage context length while preserving the full reasoning trajectory.

\Paragraph{System Prompt.}
The complete system prompt is presented in Listing~\ref{lst:claude_prompt}.

\begin{lstlisting}[
  caption={Claude Opus 4.6 system prompt.},
  label={lst:claude_prompt},
  basicstyle=\scriptsize\ttfamily,
  breaklines=true,
  frame=single,
  xleftmargin=2em,
  framexleftmargin=1.5em,
  numbers=left,
  numberstyle=\tiny\color{gray}
]
You are a computer automation agent.  You will be shown a screenshot of a
computer screen together with a task description.  Your job is to decide the
single best next action to take toward completing the task.

You MUST respond with ONLY a valid JSON object -- no markdown, no code fences,
no surrounding text of any kind.

JSON format:
{
  "Observation": "Brief description of the current screen state",
  "Reasoning": "Step-by-step reasoning that leads to the chosen action",
  "Action": "<action_name>",
  "Coordinate": [x, y],
  "Text": "text or key combination",
  "ScrollDirection": "up|down|left|right",
  "ScrollAmount": 3,
  "StartCoordinate": [x, y],
  "MilestoneCompleted": false
}

Valid actions and their required fields:
  left_click    -- Coordinate=[x,y]
  right_click   -- Coordinate=[x,y]
  double_click  -- Coordinate=[x,y]
  drag          -- StartCoordinate=[x,y], Coordinate=[x,y] (end position)
  type          -- Text="string to type"
  key           -- Text="key or combo"  (e.g. "enter", "ctrl+c")
  scroll        -- Coordinate=[x,y], ScrollDirection, ScrollAmount
  wait          -- (no extra fields needed)
  done          -- set MilestoneCompleted=true  (ONLY when the full task is complete)
  fail          -- task cannot be completed

Coordinates are in a 1280x720 pixel space.
Omit fields that are not needed for the chosen action.
Respond with ONLY the JSON object.
\end{lstlisting}

\subsubsection{Qwen3-VL (Alibaba)}
\label{sec:supp_qwen3vl}

We deploy the Qwen3-VL-32B-Thinking variant (32B parameters, open-weight) via a self-hosted vLLM inference endpoint. Adhering closely to the OSWorld prompt template~\cite{xie2024osworld}, the agent employs a \emph{tool-call XML} output format: the system prompt defines the canonical \texttt{computer\_use} function inside \texttt{<tools>} XML tags, including the standard environment description and action vocabulary, and the model is instructed to produce each action as a JSON object within \texttt{<tool\_call>...</tool\_call>} delimiters.

\Paragraph{Coordinate Convention.}
The model supports both \emph{relative} (0--999 normalised grid) and \emph{absolute} (processed image pixel) coordinate modes. In relative mode, the system prompt advertises a $1000\times1000$ virtual resolution; coordinates are converted via $x_{\text{screen}} = \lfloor x \cdot w_{\text{screen}} / 999 \rfloor$. In absolute mode, the model outputs coordinates in the processed (smart-resized) image space, which are then rescaled to the screen resolution.

\Paragraph{Image Processing.}
Screenshots are smart-resized using a factor-of-32 rounding scheme with a maximum pixel budget of $16\times16\times4\times12800$, preserving the aspect ratio while fitting within the model's vision encoder constraints.

\Paragraph{System Prompt.}
The full system prompt, including the embedded tool definition and response format specification, is presented in Listing~\ref{lst:qwen3vl_prompt}.

\begin{lstlisting}[
  caption={Qwen3-VL system prompt (tool-call XML format).},
  label={lst:qwen3vl_prompt},
  basicstyle=\scriptsize\ttfamily,
  breaklines=true,
  frame=single,
  xleftmargin=2em,
  framexleftmargin=1.5em,
  numbers=left,
  numberstyle=\tiny\color{gray}
]
# Tools

You may call one or more functions to assist with the user query.

You are provided with function signatures within <tools></tools> XML tags:
<tools>
{"type": "function", "function": {"name": "computer_use",
  "description": "Use a mouse and keyboard to interact with a computer, and
    take screenshots. [...] The screen's resolution is 1000x1000. [...]",
  "parameters": {"properties": {
    "action": {"enum": ["key","type","mouse_move","left_click",
      "left_click_drag","right_click","middle_click","double_click",
      "scroll","wait","terminate"], "type": "string"},
    "keys": {"type": "array"}, "text": {"type": "string"},
    "coordinate": {"type": "array"}, "pixels": {"type": "number"},
    "time": {"type": "number"},
    "status": {"type": "string", "enum": ["success","failure"]}
  }, "required": ["action"], "type": "object"}}}
</tools>

For each function call, return a json object with function name and arguments
within <tool_call></tool_call> XML tags:
<tool_call>
{"name": <function-name>, "arguments": <args-json-object>}
</tool_call>

# Response format
1) Action: a short imperative describing what to do in the UI.
2) A single <tool_call>...</tool_call> block.

Rules:
- Output exactly in the order: Action, <tool_call>.
- Be brief: one sentence for Action.
- If finishing, use action=terminate in the tool call.
\end{lstlisting}

\subsubsection{UITars 1.5 (ByteDance Seed)}
\label{sec:supp_uitars}

UITars 1.5 is a GUI-native vision-language model served via a self-hosted vLLM endpoint. Its prompt design inherits the OSWorld action vocabulary~\cite{xie2024osworld} (click, drag, scroll, type, hotkey, wait, finished) but re-expresses it through a Python-style function-call syntax that is native to the UITars model family. The agent supports two inference modes: a \emph{thinking} mode in which the model produces an explicit chain-of-thought enclosed in \texttt{<think>...</think>} tags before the action, and a \emph{non-thinking} mode that outputs the Thought--Action pair directly.

\Paragraph{Coordinate Convention.}
UITars outputs coordinates in a 0--1000 normalised space via \texttt{<point>x y</point>} tokens. These are converted to relative 0--1 values by the parser and subsequently multiplied by the screen resolution to yield final pixel coordinates.

\Paragraph{Action Space.}
The action vocabulary is specified through a Python-style function-call syntax (e.g., \texttt{click(point='<point>x1 y1</po\\int>')}), which is natively understood by the UITars model family. This contrasts with the JSON- or XML-based schemas used by other agents.

\Paragraph{System Prompt (Thinking Mode).}
Listing~\ref{lst:uitars_prompt} presents the system prompt used when thinking mode is enabled (\texttt{COMPUTER\_USE\_DOUBAO}).

\begin{lstlisting}[
  caption={UITars 1.5 system prompt (thinking mode).},
  label={lst:uitars_prompt},
  basicstyle=\scriptsize\ttfamily,
  breaklines=true,
  frame=single,
  xleftmargin=2em,
  framexleftmargin=1.5em,
  numbers=left,
  numberstyle=\tiny\color{gray}
]
You are a GUI agent. You are given a task and your action history, with
screenshots. You need to perform the next action to complete the task.

## Output Format
You should first think about the reasoning process in the mind and then
provide the user with the answer.
The reasoning process is enclosed within <think> </think> tags.
After the <think> tags, you should place the final answer, which concludes
your summarized thought and your action.

For example,
<think>detailed reasoning content here</think>
Thought: a small plan and finally summarize your next action in one sentence
Action: ...

## Action Space
click(point='<point>x1 y1</point>')
left_double(point='<point>x1 y1</point>')
right_single(point='<point>x1 y1</point>')
drag(start_point='<point>x1 y1</point>', end_point='<point>x2 y2</point>')
hotkey(key='ctrl c')
type(content='xxx')
scroll(point='<point>x1 y1</point>', direction='down|up|right|left')
wait()
finished(content='xxx')

## Note
- Write a small plan and summarize your next action in one sentence in Thought.
- If repeated actions have no effect, try a modified action.

## User Instruction
{instruction}
\end{lstlisting}


\subsubsection{EvoCUA}
\label{sec:supp_evocua}

EvoCUA (32B parameters, open-weight) is a native multimodal GUI agent model deployed via a self-hosted vLLM inference endpoint. Its architectural design is grounded in the OSWorld evaluation protocol~\cite{xie2024osworld} and formulated around a \textbf{Tool-Call} paradigm that directly instantiates the XML-based \texttt{computer\_use} tool-calling schema. Crucially, EvoCUA extends the canonical action vocabulary with stateful \texttt{key\_down} and \texttt{key\_up} primitives, thereby enabling fine-grained modifier-held interactions---such as Shift-constrained dragging or Alt-held scrubbing---that are indispensable for professional multimedia post-production workflows requiring precise cross-modal alignment between visual feedback and keyboard state.

\Paragraph{Prompt Architecture.}
The prompt structure is aligned with the tool-call XML paradigm: a \texttt{computer\_use} function is formally defined within \texttt{<tools>} tags, and the model orchestrates each interaction step by emitting an \texttt{Action:} line articulating the high-level intent, followed by a structured \texttt{<tool\_call>} block that encapsulates the executable operation. Specifically, the action vocabulary encompasses standard GUI primitives (\texttt{left\_click}, \texttt{right\_click}, \texttt{double\_click}, \texttt{type}, \texttt{key}, \texttt{scroll}, \texttt{mouse\_move}, \texttt{left\_click\_drag}) alongside the aforementioned stateful keyboard actions (\texttt{key\_down}, \texttt{key\_up}) and a \texttt{triple\_click} convenience action. This enriched action space is specifically designed to alleviate the expressiveness gap encountered when agents must manipulate timeline-centric, audio-visual editing interfaces where temporal synchronization between held modifier keys and spatial cursor trajectories is critical.

\Paragraph{Coordinate Convention.}
EvoCUA defaults to a $1000\times1000$ relative coordinate grid (configurable to absolute mode), which normalizes heterogeneous screen resolutions into a unified spatial representation. Screenshots are smart-resized with a factor-of-32 scheme to maintain compatibility with the vision encoder's patch-alignment requirements, ensuring robust cross-modal grounding between the visual observation and the agent's spatial reasoning.

\Paragraph{System Prompt.}
The system prompt follows the same tool-call XML structure as Qwen3-VL (Listing~\ref{lst:qwen3vl_prompt}), with the addition of \texttt{key\_down}, \texttt{key\_up}, and \texttt{triple\_click} to the action enum. The full system prompt is presented in Listing~\ref{lst:evocua_s2_prompt}.

\begin{lstlisting}[
  caption={EvoCUA system prompt (Tool-Call mode, abbreviated).},
  label={lst:evocua_s2_prompt},
  basicstyle=\scriptsize\ttfamily,
  breaklines=true,
  frame=single,
  xleftmargin=2em,
  framexleftmargin=1.5em,
  numbers=left,
  numberstyle=\tiny\color{gray}
]
# Tools
You are provided with function signatures within <tools></tools> XML tags:
<tools> {computer_use tool definition} </tools>

For each function call, return a JSON object within <tool_call></tool_call>:
<tool_call>
{"name": "computer_use", "arguments": {...}}
</tool_call>

# Response Format
1) Action: a short imperative describing what to do in the UI.
2) A single <tool_call>...</tool_call> block.
Rules: Output exactly in the order Action, <tool_call>. Be brief.

# Action Enum
key, key_down, key_up, type, mouse_move, left_click, left_click_drag,
right_click, middle_click, double_click, triple_click, scroll, wait,
terminate
\end{lstlisting}

\subsubsection{Gemini 3 Flash (Google)}
\label{sec:supp_gemini}

In contrast to the four agents described above, Gemini 3 Flash (proprietary, closed-weight; accessed via the Google API) does \emph{not} follow the OSWorld prompt template. Instead, it employs the \emph{unified planner JSON schema} developed specifically for the CutVerse autonomous pipeline. This schema outputs a structured JSON object that includes observation, reasoning, a milestone-completion flag, and an \texttt{actions} array supporting compound multi-action steps---capabilities that go beyond the single-action-per-turn paradigm of the OSWorld template.

\Paragraph{Coordinate Convention.}
The model operates in a 0--1000 normalised coordinate space where $(0, 0)$ denotes the top-left corner and $(1000, 1000)$ the bottom-right. Screenshots are resized so that the longest side does not exceed 1280 pixels (preserving aspect ratio). The normalised coordinates are converted to screen pixels by the framework's \texttt{convert\_action\_coords\_to\_screen()} utility.

\Paragraph{Multi-turn History.}
The agent maintains a sliding window of up to 5 screenshots in the conversation context. For turns within the image window, both the screenshot and a brief task reminder are included; for older turns, only a textual placeholder is retained.

\Paragraph{System Prompt.}
The system prompt defines the full action vocabulary (mouse clicks, press/release, drag, scroll, keyboard, and system actions) and specifies the JSON output schema. The complete prompt is presented in Listing~\ref{lst:gemini_prompt}.

\begin{lstlisting}[
  caption={Gemini 3 Flash system prompt.},
  label={lst:gemini_prompt},
  basicstyle=\scriptsize\ttfamily,
  breaklines=true,
  frame=single,
  xleftmargin=2em,
  framexleftmargin=1.5em,
  numbers=left,
  numberstyle=\tiny\color{gray}
]
You are an autonomous GUI automation agent for computer-use tasks on a
{os_name} device. You are operating in AUTONOMOUS MODE - you must plan and
act independently.

You must perform TWO tasks in ONE response:
1. Plan: Analyze the screenshot and decide what action to take next
2. Act: Generate the precise action(s) with exact coordinates/parameters

## Coordinate System
Output coordinates in normalized 0-1000 range:
- (0, 0) = top-left corner
- (1000, 1000) = bottom-right corner

## Action Types
### Mouse - Click: click, right_click, middle_click, double_click
### Mouse - Press/Release: mouse_down, mouse_up
### Mouse - Move/Drag/Scroll: move, drag, scroll
### Keyboard: type, key, hotkey, key_down, key_up
### System: wait, stop

## Output Format (JSON)
{
    "Observation": str,
    "Reasoning": str,
    "Action": str | null,
    "MilestoneCompleted": bool,
    "actions": [
        {
            "action_type": str,
            "x": int | null,       // 0-1000 normalized
            "y": int | null,       // 0-1000 normalized
            "text": str | null,
            "key": str | null,
            "keys": [str] | null,
            "scroll_x": int | null,
            "scroll_y": int | null,
            "end_x": int | null,
            "end_y": int | null
        }
    ]
}

## CRITICAL RULES
1. MilestoneCompleted=true AND actions=[{"action_type":"stop"}]: Goal achieved
2. MilestoneCompleted=false AND actions=[action]: Action needed
3. Consider history to avoid repeating failed attempts
4. Always output actions as an array
\end{lstlisting}

\subsubsection{Summary of Agent Configurations}
\label{sec:supp_agent_summary}

\tabref{agent_config} summarises the key configuration parameters across all five agents.
\tabref{vllm_configs} details the exact vLLM~\cite{kwon2023efficientmemorymanagementlarge}\footnote{\url{https://github.com/vllm-project/vllm}} deployment parameters utilized for the evaluated multimodal models. To ensure strict reproducibility and standardized inference across our benchmark, we explicitly outline the hardware allocation (Tensor Parallelism) and structural constraints (maximum context length and multimodal prompt limits) applied during testing. By universally deploying models through this transparent and controlled infrastructure, CutVerse guarantees that all baseline evaluations are fair, verifiable, and readily reproducible by the research community.

\begin{table}[!h]
\centering
\caption{Detailed vLLM deployment configurations, software environment, and empirical memory footprint.}
\label{tab:vllm_configs}
\small
\renewcommand{\arraystretch}{1.2}
\resizebox{\linewidth}{!}{
\begin{tabular}{l c c c c c}
\toprule
\multicolumn{6}{c}{\textbf{Global Infrastructure \& Software Environment}} \\
\midrule
\textbf{Hardware Node}     & \multicolumn{5}{l}{4 $\times$ NVIDIA RTX 5090} \\
\textbf{Software Stack}    & \multicolumn{5}{l}{CUDA 12.8.61, PyTorch 2.8.0+cu128, vLLM 0.11.0, Transformers 4.57.1} \\
\textbf{Multimodal Limits} & \multicolumn{5}{l}{Max 5 Images, 0 Videos (\texttt{--limit-mm-per-prompt})} \\
\midrule
\multicolumn{6}{c}{\textbf{Model-Specific Execution Parameters}} \\
\midrule
\textbf{Served Model} & \textbf{GPUs} & \textbf{TP Size} & \textbf{Max Context} & \textbf{GPU Util.} & \textbf{Peak VRAM} \\
\midrule
UI-TARS-1.5-7B~\cite{qin2025ui}        & 2 & 2 & 98,304 & 0.92    & 63,170 MB \\
Qwen3-VL-32B-T~\cite{yang2025qwen3technicalreport} & 4 & 4 & 28,488 & Default & 126,688 MB \\
EvoCUA-32B~\cite{xue2026evocuaevolvingcomputeruse}            & 4 & 4 & 34,768 & Default & 127,508 MB \\
\bottomrule
\end{tabular}
}
\end{table}

\begin{table}[!htbp]
  \centering
  \caption{Overall failure statistics and completion-to-execution consistency gaps. Task Consistency Gap is Task Completion Rate minus Task Execution Accuracy. Milestone Consistency Gap is Milestone Completion Rate minus Milestone Execution Accuracy.}
  \label{tab:supp_overall_failure_consistency}
  \footnotesize
  \setlength{\tabcolsep}{3pt} 
  \begin{tabular}{@{} l c c c c c c @{}}
  \toprule
  Model 
  & \makecell{Incomplete\\Tasks} 
  & \makecell{Incomplete\\Task Ratio} 
  & \makecell{Incomplete\\Milestones} 
  & \makecell{Incomplete\\Milestone Ratio} 
  & \makecell{Task\\Consistency Gap} 
  & \makecell{Milestone\\Consistency Gap} \\
  \midrule
  Claude Opus 4.6~\cite{anthropic2026claude46}      & 59/186  & 0.317 & 159/631 & 0.252 & 0.116 & 0.016 \\
  Gemini 3 Flash~\cite{gemini3_2026}                & 61/186  & 0.328 & 187/631 & 0.296 & 0.091 & 0.015 \\
  EvoCUA-32B~\cite{xue2026evocuaevolvingcomputeruse}& 90/186  & 0.484 & 283/631 & 0.448 & 0.116 & 0.012 \\
  Qwen3-VL-32B-T~\cite{yang2025qwen3technicalreport}& 96/186  & 0.516 & 295/631 & 0.468 & 0.148 & 0.015 \\
  UITars-1.5-7B~\cite{qin2025ui}                    & 104/186 & 0.559 & 314/631 & 0.498 & 0.123 & 0.014 \\
  \bottomrule
  \end{tabular}
\end{table}

\section{Additional Experimental Results}
\label{sec:Additional}
\subsection{Additional Data Statistics}

\begin{table}[!htbp]
  \centering
  \caption{Task-type action distribution profile.}
  \label{tab:tasktype_actions_success}
  \small
  \setlength{\tabcolsep}{4pt} 
  \begin{tabular}{@{} l c c c @{}} 
  \toprule
  Task Type & Top-1 & Top-2 & Top-3 \\
  \midrule
  Masking, Matting, and Tracking   & Left click (48.0\%) & Drag (24.8\%)       & Key press (17.3\%) \\
  \addlinespace
  Effects and Visual Tuning        & Left click (59.4\%) & Drag (19.7\%)       & Key press (6.7\%) \\
  \addlinespace
  Audio and Rhythm Editing         & Left click (44.3\%) & Key press (30.6\%)  & Drag (11.4\%) \\
  \addlinespace
  Timeline Editing and Arrangement & Key press (36.2\%)  & Left click (35.9\%) & Drag (18.5\%) \\
  \addlinespace
  Asset Import and Management      & Left click (44.0\%) & Key press (30.8\%)  & Drag (18.2\%) \\
  \addlinespace
  Preview, Check, and Validation   & Left click (71.4\%) & Drag (14.3\%)       & Wait (9.1\%) \\
  \addlinespace
  Export and Delivery              & Left click (53.9\%) & Key press (21.7\%)  & Drag (8.1\%) \\
  \addlinespace
  Launch and Setup                 & Left click (69.1\%) & Key press (8.8\%)   & Wait (7.4\%) \\
  \addlinespace
  Generative Workflow              & Left click (55.0\%) & Drag (17.5\%)       & Key press (11.2\%) \\
  \bottomrule
  \end{tabular}
\end{table}

\begin{table*}[!t]
\centering
\caption{Detailed breakdown of dominant task types and interaction modalities across different software environments.}
\label{tab:software_tasks_interactions}
\small
\setlength{\tabcolsep}{6pt}
\begin{tabular}{l p{4.3cm} p{4.3cm} p{4.3cm}}
\toprule
Software & Top-1 & Top-2 & Top-3 \\
\midrule
\multicolumn{4}{c}{\textbf{Part A: Dominant Task Types}} \\
\midrule
After Effects   & Effects and visual tuning (61.5\%) & Export and delivery (19.2\%) & Masking and tracking (7.7\%) \\
\addlinespace
ComfyUI         & Generative workflow (50.0\%) & Export and delivery (33.3\%) & Asset import and management (16.7\%) \\
\addlinespace
DaVinci Resolve & Effects and visual tuning (50.0\%) & Export and delivery (20.0\%) & Timeline editing (15.0\%) \\
\addlinespace
JianYing        & Effects and visual tuning (18.8\%) & Export and delivery (18.8\%) & Asset management (15.9\%) \\
\addlinespace
Jimeng          & Effects and visual tuning (20.0\%) & Audio editing (16.0\%) & Export and delivery (16.0\%) \\
\addlinespace
Keling          & Asset management (26.9\%) & Preview and validation (15.4\%) & Export and delivery (11.5\%) \\
\addlinespace
Premiere Pro    & Audio editing (25.9\%) & Effects tuning (20.4\%) & Asset management (13.0\%) \\
\addlinespace
Photoshop       & Effects and visual tuning (24.2\%) & Export and delivery (18.2\%) & Asset management (12.1\%) \\
\midrule
\multicolumn{4}{c}{\textbf{Part B: Dominant Interaction Modalities}} \\
\midrule
After Effects   & Left click (62.1\%) & Drag (24.4\%) & Scroll (4.7\%) \\
ComfyUI         & Left click (53.2\%) & Drag (21.0\%) & Scroll (6.5\%) \\
DaVinci Resolve & Left click (48.5\%) & Drag (24.4\%) & Key press (7.8\%) \\
JianYing        & Left click (58.9\%) & Drag (17.1\%) & Key press (11.3\%) \\
Jimeng          & Left click (58.6\%) & Key press (22.4\%) & Drag (8.9\%) \\
Keling          & Left click (71.3\%) & Drag (18.1\%) & Type text (3.2\%) \\
Premiere Pro    & Left click (43.7\%) & Key press (31.8\%) & Drag (12.1\%) \\
Photoshop       & Left click (46.0\%) & Key press (31.2\%) & Drag (11.3\%) \\
\bottomrule
\end{tabular}
\end{table*}

\begin{table}[!t]
\centering
\caption{The formalized nine-category taxonomy of media post-production tasks in CutVerse. Each category is grounded in authentic task instructions and evaluated via granular multimodal QA pairs to rigorously verify spatial-temporal state transitions.}
\label{tab:task_taxonomy_examples}
\small
\renewcommand{\arraystretch}{1.5} 
\begin{tabular}{p{3.8cm} p{5.5cm} p{6.2cm}}
\toprule
\textbf{Task Category} & \textbf{Authentic Task Instruction} & \textbf{Multimodal QA Evaluation Example} \\
\midrule

\textbf{Effects and visual tuning} 
& \textit{Applying and Adjusting Glow Effect in DaVinci Resolve} 
& \textbf{Q:} In the right panel's 'Glow' effect settings, is the 'Input Alpha' slider value changed from 0.543 to 0.612, as observed in the slider display? \newline \textbf{A:} True \\
\midrule

\textbf{Export and delivery} 
& \textit{Exporting and Recording Workflow in Adobe Premiere Pro} 
& \textbf{Q:} Is the blue Export button clicked at the bottom right of the export settings screen with a progress dialog appearing in the center showing export progress for 'hello world'? \newline \textbf{A:} True \\
\midrule

\textbf{Asset import and management} 
& \textit{Import and Manage Video Clip in Premiere Pro} 
& \textbf{Q:} Is 'AntiguaArchTL' video thumbnail bordered in blue and marked with a checkmark in the central footage grid panel after selection? \newline \textbf{A:} True \\
\midrule

\textbf{Audio and rhythm editing} 
& \textit{Audio Editing in Adobe Premiere Pro} 
& \textbf{Q:} Does the right edge of the first audio clip extend visually to fill the gap within the bottom timeline panel after the second clip's audio is deleted? \newline \textbf{A:} True \\
\midrule

\textbf{Timeline editing and arrangement} 
& \textit{Trimming Video Clips in Adobe Premiere Pro} 
& \textbf{Q:} Is the second video clip shortened on the timeline after clicking and dragging its right edge, with a tooltip displaying the new duration above the timeline? \newline \textbf{A:} True \\
\midrule

\textbf{Preview, check, and validation} 
& \textit{Video Editing and Documentation Process} 
& \textbf{Q:} Is the playhead advancing from the position 00:00:01:00 in the bottom timeline panel, with the audio meters below the preview window showing activity? \newline \textbf{A:} True \\
\midrule

\textbf{Masking, matting, and tracking} 
& \textit{Mask Animation Creation in Adobe After Effects} 
& \textbf{Q:} Is the colorful image fully revealed in the center composition panel, with the mask path expanded to cover the entire area? \newline \textbf{A:} True \\
\midrule

\textbf{Launch and setup} 
& \textit{Transitioning Between Applications and Opening Projects} 
& \textbf{Q:} Has the Adobe Premiere Pro interface switched to the main editing workspace after selecting the 'task1' project, displaying the media browser, preview monitor, and timeline? \newline \textbf{A:} True \\
\midrule

\textbf{Generative workflow} 
& \textit{Semantic Guidance Input for Video Processing Workflow} 
& \textbf{Q:} Is the 'positive\_prompt' input field in the 'WanVideo T5 text encoder' node filled with Chinese text? \newline \textbf{A:} True \\
\bottomrule
\end{tabular}
\end{table}

\Paragraph{Task Distribution Heterogeneity.} 
Part A of \tabref{software_tasks_interactions} details the distribution of dominant tasks, exposing the profound heterogeneity of media workflows. While \textit{Effects and visual tuning} predominantly leads in visually intensive tools such as After Effects (61.5\%) and DaVinci Resolve (50.0\%), \textit{Export and delivery} serves as a universally mandatory concluding milestone across nearly all platforms. Furthermore, the distribution captures specific functional biases, highlighting ComfyUI's reliance on \textit{Generative workflows} (50.0\%) and Premiere Pro's emphasis on \textit{Audio editing} (25.9\%). Crucially, varying degrees of task concentration directly dictate execution complexity. Professional environments such as After Effects and ComfyUI exhibit highly concentrated distributions (Top-1 task $\ge$ 50\%), demanding deep, long-horizon specialized operations within a single domain. Conversely, comprehensive editing platforms such as Premiere Pro and Photoshop display a flattened distribution (Top-1 hovering around 20-25\%), necessitating rapid context-switching between diverse modalities, thereby imposing stricter demands on sustained cross-modal reasoning.

\Paragraph{Interaction Modality Complexity.} 
Part B of \tabref{software_tasks_interactions} delineates the execution modalities, exposing the mechanical bottlenecks of professional environments. While \textit{Left click} predictably serves as the foundational operation across all platforms, peaking at 71.3\% in web-centric tools such as Keling, the true complexity emerges in secondary interactions. Professional software, particularly Premiere Pro and Photoshop, exhibits a substantial reliance on \textit{Key press} operations (31.8\% and 31.2\%, respectively). This underscores the absolute necessity of keyboard shortcuts for rapid tool switching and global viewport navigation. Furthermore, continuous \textit{Drag} operations manifest as a universal requirement for precise spatial and temporal adjustments, notably in After Effects (24.4\%) and DaVinci Resolve (24.4\%). These distributions firmly establish that basic point-and-click capabilities are fundamentally inadequate; mastering tightly coupled, compositional key-mouse coordination and sustained dragging is strictly mandatory for autonomous agents in creative workflows.

\Paragraph{Task-Driven Action Mapping.} 
\tabref{tasktype_actions_success} delineates the intrinsic correlation between post-production task typologies and their underlying execution actions, revealing that an agent's required interaction modality is strictly dictated by the specific task type. While foundational \textit{Left click} actions predictably dominate procedurally driven tasks such as \textit{Preview, check, and validation} (71.4\%) and \textit{Launch and setup} (69.1\%), structurally complex tasks demand entirely distinct action mappings. Specifically, \textit{Timeline editing and arrangement} exhibits a fundamental shift towards \textit{Key press} operations (36.2\%), highlighting a direct relationship between structural video assembly and the absolute necessity of keyboard shortcuts for rapid tool switching. Furthermore, tasks requiring precise spatial-temporal grounding, such as \textit{Masking, matting, and tracking} and \textit{Effects and visual tuning}, establish a strong dependency on continuous \textit{Drag} actions (24.8\% and 19.7\%, respectively) for manipulating bezier paths and parameter sliders. Conversely, organizational tasks such as \textit{Asset import and management} tightly couple with \textit{Key press} inputs (30.8\%) to facilitate semantic search and file routing. Ultimately, this distribution confirms that the execution action space is fundamentally task-dependent; succeeding in diverse multimedia workflows requires agents to dynamically map specific task semantics to tightly coupled, compositional key-mouse action combinations.


\subsection{Additional Evaluation Analysis}
\Paragraph{Failure Analytics and Execution Consistency.} 
Table~\ref{tab:supp_overall_failure_consistency} details the overall failure statistics and consistency metrics across the evaluated multimodal agents, exposing a profound gap between current capabilities and professional production standards. Even industry-leading models struggle significantly; Claude and Gemini exhibit incomplete task ratios of 31.7\% and 32.8\%, respectively. Meanwhile, open-weight and UI-centric models, specifically Qwen and UI-TARS, fail to complete over half of the assigned tasks (51.6\% and 55.9\%). 

Beyond absolute failure rates, the data reveals a critical vulnerability in standard agent evaluation paradigms through the "Consistency Gap"—the mathematical difference between the perceived completion rate and the strictly verified execution accuracy. At the macro-task level, agents demonstrate a severe task consistency gap, peaking at 0.148 for Qwen and 0.116 for Claude. This significant discrepancy indicates that agents frequently suffer from execution hallucinations; they declare task completion despite failing to achieve the precise spatial-temporal requirements of the creative workflow. Conversely, the milestone consistency gap remains exceptionally tight across all models, strictly bounded between 0.012 and 0.016. This striking contrast definitively proves the necessity of our granular evaluation infrastructure. By decomposing continuous workflows into visually grounded, intermediate milestones, CutVerse effectively eliminates false positives, ensuring that an agent's perceived progress perfectly aligns with its authentic execution accuracy in complex multimodal environments.

\clearpage
\bibliographystyle{plainnat}
\bibliography{main}

\end{document}